\pdfoutput=1 
\documentclass{article}
\usepackage{microtype}
\usepackage{float}
\usepackage{graphicx}
\usepackage{subcaption}
\usepackage{booktabs}
\usepackage{caption}
\usepackage{multirow}
\usepackage{hyperref}
\usepackage{tabularx}
\usepackage{amsmath}
\usepackage{amssymb}
\usepackage{mathtools}
\usepackage{amsthm}
\usepackage{tikz}
\usepackage{lineno}
\usepackage{xcolor}
\usetikzlibrary{calc}
\usepackage{pgfplots}
\pgfplotsset{compat=1.18}

\usepackage{iclr2026_conference,times}

\title{The Capability Frontier: Benchmarks Miss 82\% of Model Performance}

\iclrfinalcopy

\author{%
\parbox{0.98\textwidth}{\centering
Bradley Fowler\textsuperscript{1}, Ryan Smith\textsuperscript{1}, Daniel Thi Graviet\textsuperscript{1},
William Myers\textsuperscript{1}, Joshua Greaves\textsuperscript{1}, Narmeen Fatimah Oozeer\textsuperscript{1},
Antía García\textsuperscript{1}, Philip Quirke\textsuperscript{1}, Amirali Abdullah\textsuperscript{3,1},
Fazl Barez\textsuperscript{1,2}, Shriyash Kaustubh Upadhyay\textsuperscript{1}\\[0.6em]
{\normalfont\textsuperscript{1}Martian \qquad \textsuperscript{2}University of Oxford \qquad \textsuperscript{3}ThoughtWorks}}%
}

\begin{document}
\maketitle

\begin{abstract}
Existing benchmarks typically report accuracy for a single model on a single run. This systematically understates real-world LLM capabilities, particularly under heterogeneous data distributions: (i) different models get different questions correct according to their specializations, and (ii) given a budget, multiple generations can be sampled and selectively retained. To quantify this gap, we introduce the \textit{Capability Frontier}: a Pareto frontier over a set of models that characterizes the best achievable performance at each cost level under optimal selection across models and generations (i.e., via an oracle).
Our construction corrects for two opposing biases: underestimation from single-model evaluation and overestimation from taking maxima over noisy samples. We study 21 LLMs across 16 widely used benchmarks spanning coding, reasoning, medicine, factuality, instruction following, and agentic tasks, comparing Capability Frontier performance at matched cost to each benchmark’s top-performing model. Correcting for single-model evaluation yields a 54\% error rate reduction; additionally correcting for single runs yields an 82\% improvement, with SOTA accuracy matched at 85\% cost reduction. Complementing these empirical results, we use controlled probabilistic simulations to show that higher query topic entropy produces a near-monotonic increase in the performance gap between oracle routing and the best single model. Our findings suggest collective LLM capabilities are substantially underestimated, with implications for evaluation and deployment in data-heterogeneous, multi-domain settings.

\end{abstract}



\section{Introduction}
\label{sec:intro}

LLMs in the wild face messy and disparate workloads. Consider the example of medical question answering systems, where life-and-death queries form a polyphonic mixture spanning diverse subdomains of medical knowledge such as genomic variation, and human structural physiology. Consistent with this heterogeneity, \citet{singhal2025toward} show that models excel in different medical topics: GPT-4-base~\citep{achiam2023gpt} outperforms Med-PaLM-2~\citep{singhal2025toward} on MMLU Medical Genetics (97.0\% vs.\ 92.0\%) and Anatomy (85.2\% vs.\ 84.4\%), while Med-PaLM-2 excels on Professional Medicine (95.2\% vs.\ 93.8\%) and College Medicine (83.2\% vs.\ 80.9\%). An oracle selector with access to per-query ground-truth correctness could therefore outperform both models, yet this achievable performance remains unmeasured in standard evaluation.

Foundational work on LLM routing has begun to probe this gap. \citet{shnitzer2023largelanguagemodelrouting} showed that an oracle router can achieve approximately 20\% performance gains by switching models per prompt. RouterBench \citep{hu2024routerbenchbenchmarkmultillmrouting} quantified model complementarity, finding that secondary models provide unique correct answers on 10--30\% of prompts. RouteLLM \citep{ong2025routellmlearningroutellms} further demonstrated up to 2$\times$ cost reductions by identifying prompts where cheaper models suffice.
These studies estimate oracle performance from finite samples of generations per prompt, selecting the highest-performing model accordingly. Because oracle selection takes a maximum over noisy performance estimates, such procedures are positively biased, systematically overstating achievable gains. This effect is amplified under realistic generation budgets, where only limited samples ($G \leq 10$) are available.

\begin{figure*}[t]
    \centering
    \includegraphics[width=0.9\textwidth]{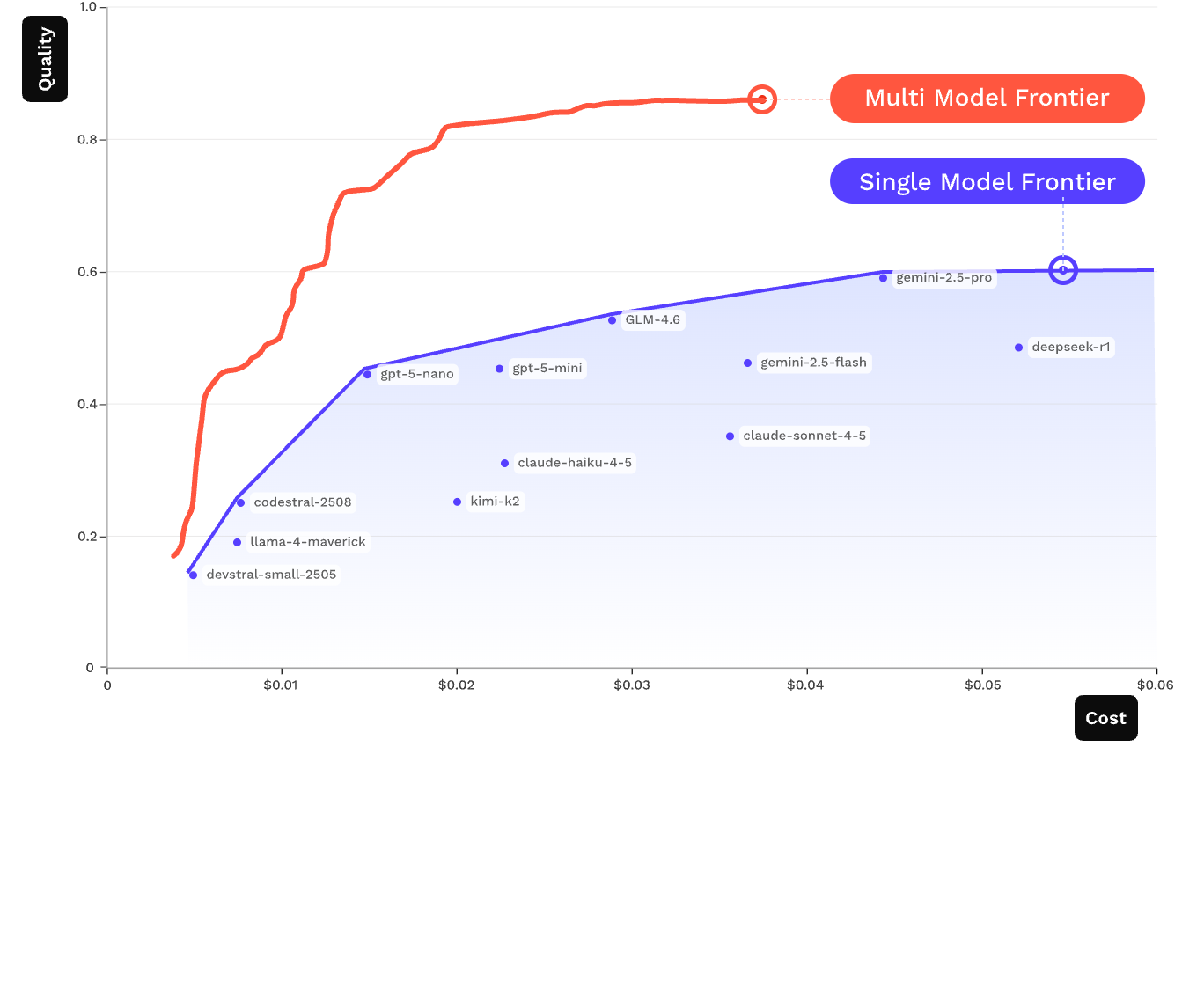}
    \caption{\textbf{The Capability Frontier:} Dynamic per-prompt LLM selection substantially outperforms any fixed LLM across our 16 benchmarks. Sample datapoints from App.~\ref{app:benchmarks} are shown. For any given cost budget, substantial quality improvements can be realized relative to a single LLM. Conversely, for a fixed quality threshold substantial cost savings can be realized through dynamic LLM selection.}
    \label{fig:fig1}
\end{figure*}

To robustify Capability Frontier estimation  and correct for these finite-sample effects, we develop debiasing methods that recover accurate frontier measurements and provide principled upper bounds on achievable performance. Our empirical analysis spans 21 LLMs across 16 benchmarks covering coding, reasoning, medicine, factuality, instruction following, and agentic tasks. The results reveal that standard single-model evaluation substantially understates achievable performance: at matched cost, the Capability Frontier achieves 54\% average error reduction compared to each benchmark's top model. When additionally accounting for multi-run selection (posthoc routing), error reduction reaches 82\%. Conversely, SOTA accuracy can be matched at 85\% lower cost on the frontier. These gaps are not merely theoretical - they represent performance that is achievable today with existing models and straightforward inference-time strategies.

Accurately measuring these gains requires care. The standard oracle computation selects the model with the highest sample mean for each prompt, then reports this mean as achievable performance. Because this takes a maximum over noisy estimates, it preferentially captures positive outliers. With limited generations per prompt ($G \leq 10$ due to cost), this bias is substantial: we find it inflates accuracy estimates by up to 8.7\% and cost estimates by up to 88\% (Sec.~\ref{ssec:bias_quant}). Our debiasing methods - namely extrapolation-based correction and probabilistic graphical modeling— enable accurate frontier estimation. Complementing these empirical findings, we construct controlled probabilistic simulations showing that oracle gains scale monotonically with workload diversity. 
To highlight our main contributions:
\begin{enumerate}
    \item \textbf{Capability Frontier.} We introduce a rigorous framework for quantifying the gap between single-model benchmark evaluation and achievable system-level performance under optimal model and generation selection.

    \item \textbf{Debiasing methods.} We show that finite-sample oracle estimators are positively biased and propose correction methods based on extrapolation and probabilistic modeling, with explicit assumptions and empirical validation.
    
    \item \textbf{Bias characterization.} We formally analyze oracle bias, showing it decays as $O(G^{-\lambda})$ with generations per prompt, and empirically validate this scaling across benchmarks.
    
    \item \textbf{Empirical evaluation.} Across 21 LLMs and 16 benchmarks, we quantify both achievable frontier gains and the magnitude of bias in naive oracle estimates. See Figure \ref{fig:fig1}.
    
    \item \textbf{Controlled simulations.} When simulating synthetic workloads spanning low to high-diversity regimes, we find that oracle gains grow monotonically with workload entropy. These results provide mechanistic grounding for oracle performance, demonstrating that achievable gains are fundamentally driven by data heterogeneity.
\end{enumerate}

\section{Related Work}
\label{sec:related}

The rapid proliferation of LLMs has increased research interest in LLM routing, the dynamic selection of models to balance quality, cost, and latency. \citet{shnitzer2023largelanguagemodelrouting} first formalized this problem using benchmark datasets, introducing the oracle router as a theoretical upper bound for performance gains. While they identified significant headroom beyond the ``best-on-average" model, their oracle relied on biased sample means, a limitation our work addresses. Subsequent frameworks like RouterBench \citep{hu2024routerbenchbenchmarkmultillmrouting} have standardized evaluation across routing methods, though they similarly utilize these biased estimates.

\paragraph{Universal and Zero-Shot Routing}
Recent methods seek to solve the ``model lock-in" problem, where routers must be retrained whenever the model pool changes. UniRoute \citep{jitkrittum2025uniroute} addresses this by representing LLMs as feature vectors based on anchor prompts, allowing for generalization to unseen models. Similarly, ZeroRouter \citep{yan2026zerorouter} utilizes a universal latent space to decouple query difficulty from specific model profiles, enabling zero-shot selection across evolving model ecosystems.

\paragraph{Theoretical Foundations}
While the industry moves toward expert orchestration for safer and more capable systems \citep{quirke2025monolithsexpertorchestrationcapable}, a gap remains between implementable routers and theoretical optimality. Our work builds upon the foundations of oracle routing \citep{shnitzer2023largelanguagemodelrouting, hu2024routerbenchbenchmarkmultillmrouting}, but departs from them by correcting for the ``optimizer's curse"—a statistical bias well-documented in economics \citep{andrewsKitagawaMccloskey2024inferenceWinners, capenClappCampbell1971competitive} and decision analysis \citep{smithWinkler2006optimizersCurse}. By introducing debiased oracles, we provide a more rigorous framework for quantifying the true headroom available in the Capability Frontier.

See Appendix \ref{app:more_routing} for more routing methods, including training-free, cascades and preference routing.

\section{Problem Setting}
\label{sec:problem}

Let $n \in [N]$ index dataset prompts, $l \in [L]$ index LLMs, and $g \in [G]$ index independent stochastic generations. For each prompt-model pair, we observe $G$ generations and evaluate each using metric $\phi_{nlg} \in \mathbb{R}$ (e.g., correctness, cost, latency).

The standard formulation for routing is a two-dimensional objective that maximizes quality whilst minimizing cost:
\begin{equation}\label{eqn:objective}
    \phi_{nlg} = \left\{ (\textbf{Q}, -\textbf{C}) \right\}_{nlg}
\end{equation}
\(\mathbf{Q}\), \(\mathbf{C}\), and \(\mathbf{T}^{95}\) are all tensors with identical dimensionality that represent Quality, Cost, and P95 latency.

\paragraph{The routing problem.}
A router $\pi: \mathcal{X} \to [L]$ maps each prompt to a model. The goal is to find $\pi$ maximizing expected performance:
\begin{equation}
    \max_\pi \frac{1}{N} \sum_n \mathbb{E}[\phi_{n, \pi(x_n), g}]
\end{equation}

\paragraph{The oracle router.}
An oracle router has access to true expected performance $\mu_{nl} = \mathbb{E}[\phi_{nlg}]$ and selects optimally:
\begin{equation}
    l^*(n) = \arg\max_l \mu_{nl}
\end{equation}
The \textbf{true oracle value} is:
\begin{equation}\label{eqn:true_oracle}
    \mathcal{O}^{true} = \frac{1}{N} \sum_n \max_l \mu_{nl}
\end{equation}
This is the fundamental upper bound for routing: the best achievable performance given perfect knowledge of each model's expected performance on each prompt.

\paragraph{The estimation problem.}
We cannot observe $\mu_{nl}$ directly, only noisy realizations $\phi_{nlg}$. The standard approach estimates $\mu_{nl}$ with the sample mean $\bar{\phi}_{nl} = \frac{1}{G}\sum_g \phi_{nlg}$ and computes:
\begin{equation}\label{eqn:biased_oracle}
    \mathcal{O}^{biased} = \frac{1}{N} \sum_n \max_l \bar{\phi}_{nl}
\end{equation}
We show next this estimator is positively biased: $\mathbb{E}[\mathcal{O}^{biased}] > \mathcal{O}^{true}$.

\section{Oracle Bias and Debiasing Methods}
\label{sec:method}

\subsection{Characterizing Oracle Bias.}
\noindent \textbf{Why the biased oracle is biased.}
The bias arises because taking the maximum over sample means preferentially selects models whose samples exceeded their true means. The bias crops up in many fields from Economics \citet{andrewsKitagawaMccloskey2024inferenceWinners}, to Management \citet{smithWinkler2006optimizersCurse}; however, it was first spotted in Auctions by \citet{capenClappCampbell1971competitive}.
This paper formalizes the bias in LLM Routing, presenting new methods to remove this bias. We formalize this under two distributional assumptions.

\subsection{Gaussian case}
Assume $\phi_{nlg} \sim \mathcal{N}(\mu_{nl}, \sigma^2_{nl})$ independently. The sample mean satisfies $\bar{\phi}_{nl} \sim \mathcal{N}(\mu_{nl}, \sigma^2_{nl}/G)$. 

To derive the bias in closed form, we make a simplifying assumption:
\begin{equation}\label{eqn:equal_means}
    \mu_{nl} = \mu_n, \quad \sigma^2_{nl} = \sigma^2_n \quad \forall l
\end{equation}

\textbf{Remark.} Assumption~\eqref{eqn:equal_means} is used \textit{only} to derive the functional form of bias decay, not to claim that $\mathcal{O}^{true} = \bar{\mu}$. Under heterogeneous means, the true oracle remains $\frac{1}{N}\sum_n \max_l \mu_{nl}$, which our debiasing methods estimate without requiring equal means.

Under~\eqref{eqn:equal_means}, the expected maximum of $L$ i.i.d. Gaussians with variance $\sigma^2_n/G$ is approximately:
\begin{equation}
    \mathbb{E}[\max_l \bar{\phi}_{nl}] \approx \mu_n + \sigma_n \sqrt{\frac{2 \log L}{G}}
\end{equation}

Averaging over prompts:
\begin{equation}\label{eqn:bias_formula}
    \mathcal{O}^{biased} \approx \underbrace{\bar{\mu}}_{\text{True Oracle}} + \underbrace{\bar{\sigma} \sqrt{\frac{2 \log L}{G}}}_{\text{Bias}}
\end{equation}
where $\bar{\mu} = \frac{1}{N}\sum_n \mu_n$ and $\bar{\sigma} = \frac{1}{N}\sum_n \sigma_n$.

\textbf{Key insight:} The bias reduces as $O(G^{-0.5})$ and increases with $L$ (more models) and $\bar{\sigma}$ (higher variance). For $G=10$ and $L=21$, this bias is non-negligible.

\subsubsection{Bernoulli case}
For binary metrics (correct/incorrect), let $\phi_{nlg} \sim \text{Bernoulli}(p_{nl})$. Under the simplifying assumption $p_{nl} = p_n$:
\begin{align}
    Y_{nl} &= \sum_g \phi_{nlg} \sim \text{Binomial}(G, p_n) \\
    \mathbb{E}[\max_l Y_{nl}] &= \dfrac{1}{NG} \sum_{n,g} \left[1 - F(g; p_n)^L\right]
\end{align}
where $F(g; p_n)$ is the Binomial CDF.

\begin{figure}[t]
\centering
\begin{subfigure}{\linewidth}
  \centering
  \includegraphics[width=0.9\linewidth]{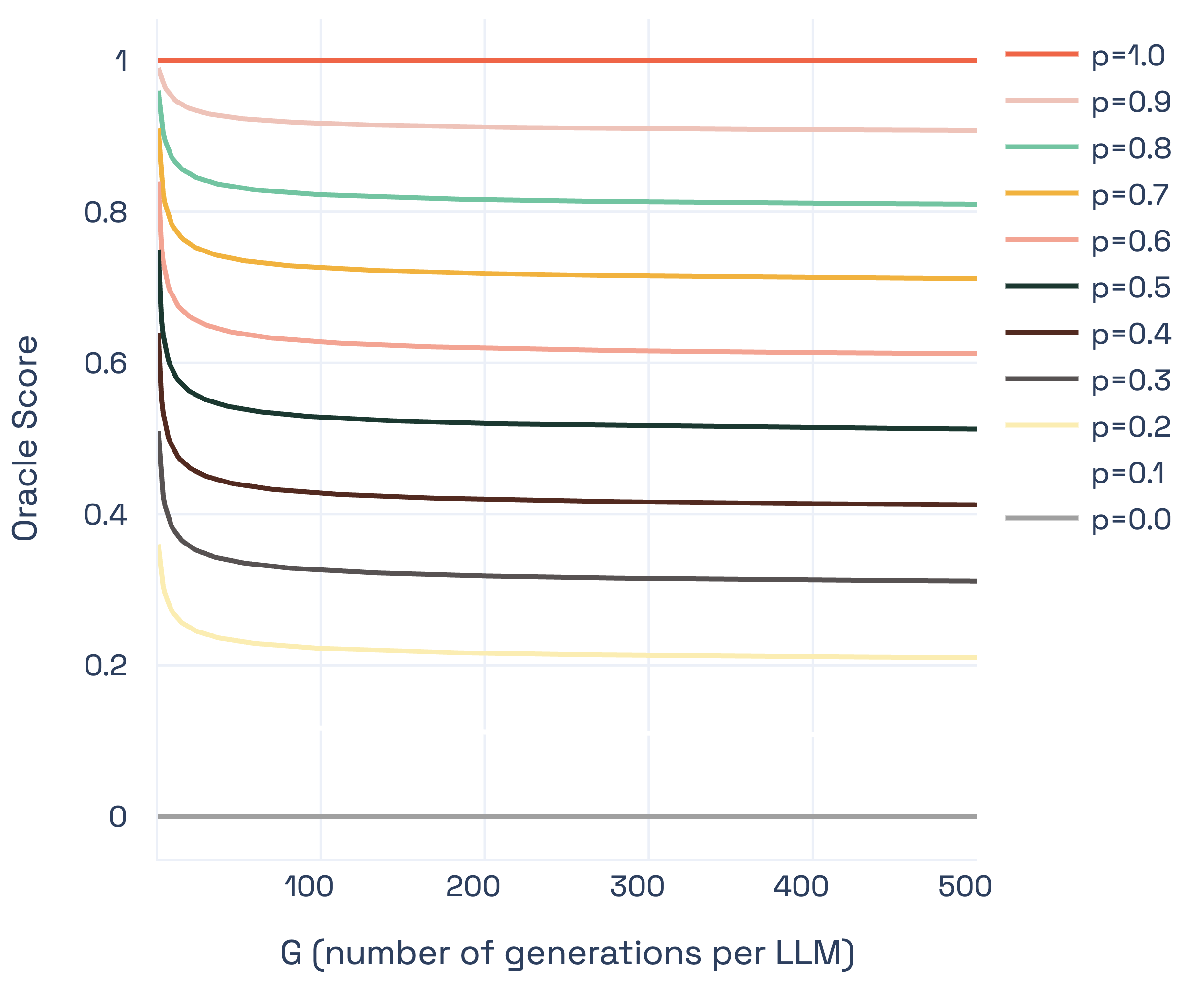}
  \caption{$L=2$ LLMs}
\end{subfigure}
\vspace{0.3em}
\begin{subfigure}{\linewidth}
  \centering
  \includegraphics[width=0.9\linewidth]{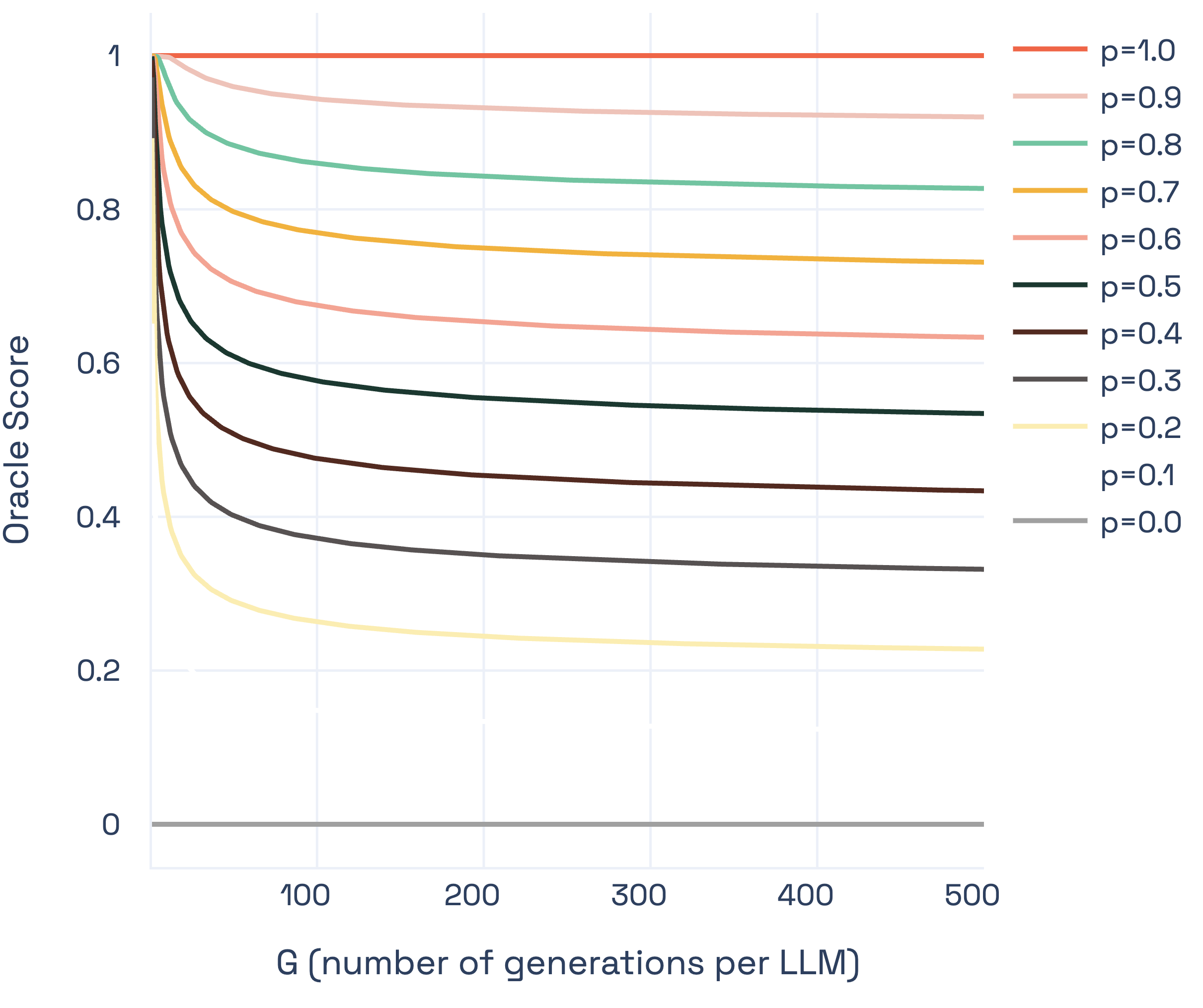}
  \caption{$L=10$ LLMs}
  \label{sfig:bias-decay}
\end{subfigure}
\caption{\textbf{Oracle bias reduces with generations.} Oracle bias is greatest when each LLM is prompted once $(G=1)$, and tends towards zero as $G$ grows. Oracle bias decays with $O(G^{-0.5})$ in the limit. Curves show LLM success rates, $p$. Always correct/incorrect LLMs ($p=0, 1$) have no bias and curves are horizontal.}
\label{fig:bias-decay}
\end{figure}

There is no clean separation of true oracle and bias term, however through empirical study we can determine the characteristics of the bias decay. We know that for large G, the Oracle should tend towards $p_n$. Fig.\ref{fig:bias-decay} shows how the bias decays in different scenarios. When $p=0$ or $p=1$, there is no variance in LLM performance per data point and so, the bias is zero for all G.

Through a synthetic study (Appendix~\ref{app:decay}), we found the bias decayed with $O(G^{-0.5})\ \forall\ L>1,\ p\in (0, 1)$ in the limit of large G for heterogeneous $(\mu_{nl}, \sigma_{nl})$ generations across models, consistent with the Gaussian analysis. For correlations generations between models, we found the exponent varied in the range $[0.25, 0.75]$ for sensible hyper-parameters. Both heterogeneous and correlated scenarios required roughly $G=50$ generations in order to fit Eqn.\ref{eqn:extrapolate} accurately.

\textbf{Key insight:} The bias reduces as $O(G^{-\lambda})$ where $\lambda \in [0.25, 0.75]$. At least $G>=50$ generations are needed for $O(G^{-\lambda})$ to be the dominate term in the bias decay.


\subsection{Debiasing Methods}
\label{ssec:debiasing}



\subsubsection{Method 1: Extrapolation}
\label{sssec:extrapolation}

Given that bias decays as $O(G^{-\lambda})$ where $\lambda \in [0.25, 0.75]$, we fit:
\begin{equation}\label{eqn:extrapolate}
    \mathcal{O}^{biased}(G) = \alpha + \beta G^{-\lambda}
\end{equation}
and estimate $\mathcal{O}^{true} = \alpha$.

\begin{figure}[t]
\centering
\includegraphics[width=0.9\linewidth]{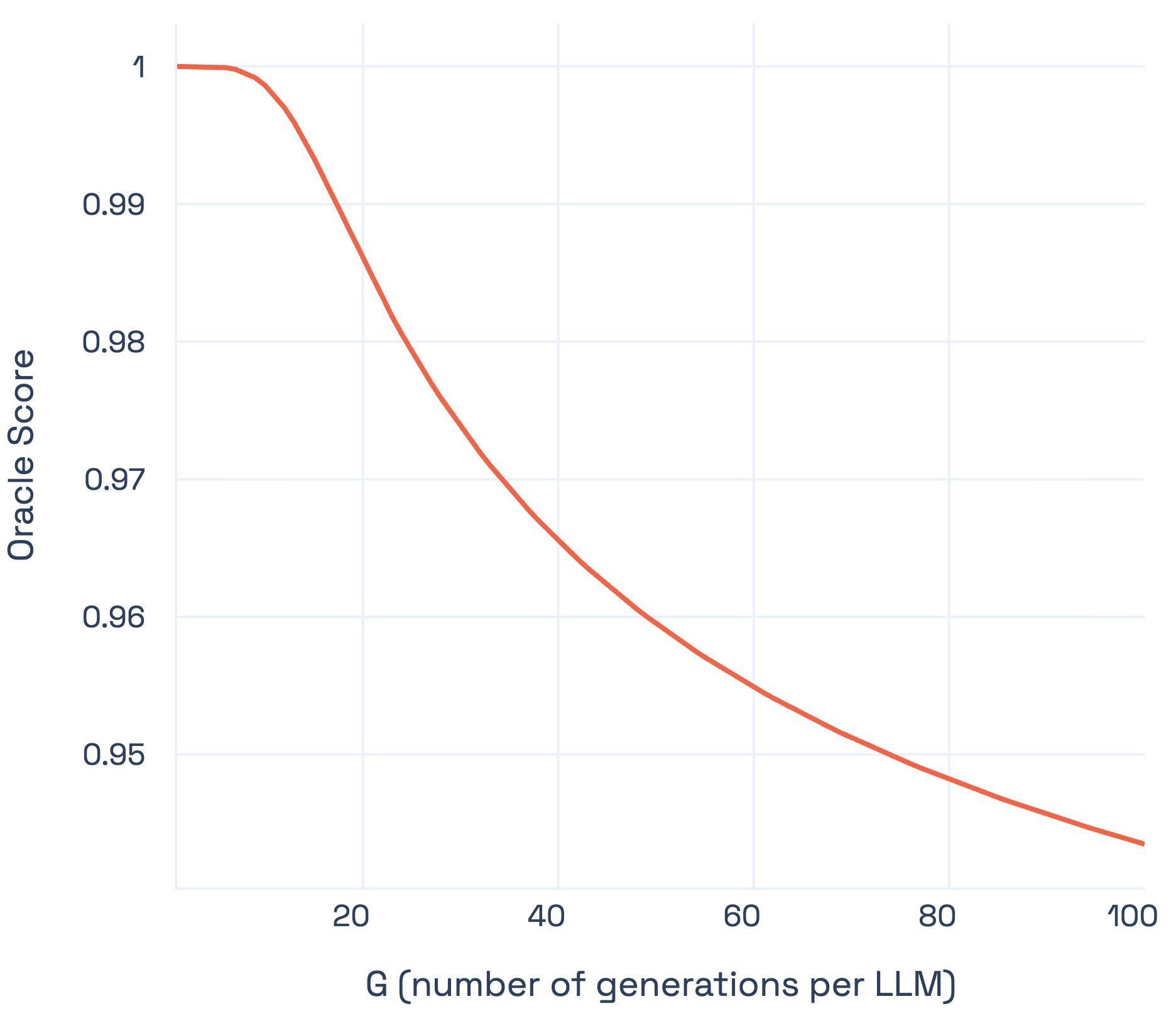}
\caption{\textbf{Bias decay deviates from $O(G^{-0.5})$ for small $G$.} At $p=0.9$, the curve only follows the asymptotic form for $G > 20$, motivating our smooth transition formulation. This curve is closeup of $p=0.9$ from Fig.~\ref{sfig:bias-decay}}.
\label{fig:curve-start}
\end{figure}

In practice, due to cost constraints we are not in the regime of $G\ge50$ and Equation \ref{eqn:extrapolate} does not hold as can be seen in Fig.~\ref{fig:curve-start}\textbf{}.
As such, a smooth transition formulation can be used to better approximate the bias decay:
\begin{equation}\label{eqn:extrapolate_smooth}
    \mathcal{O}^{biased}(G) = \alpha + \beta \left[1 + \left(\frac{G - \gamma}{\delta}\right)^2\right]^{-\lambda / 2}
\end{equation}

\textbf{Limitations.} With $G < 10$, extrapolation carries risk. We validate by: (1) testing on synthetic data with known ground truth, and (2) comparing to PGM estimates.


\subsubsection{Method 2: Probabilistic Graphical Model}
\label{sssec:pgm}

\begin{figure}[t]
    \centering
    \includegraphics[width=\linewidth]{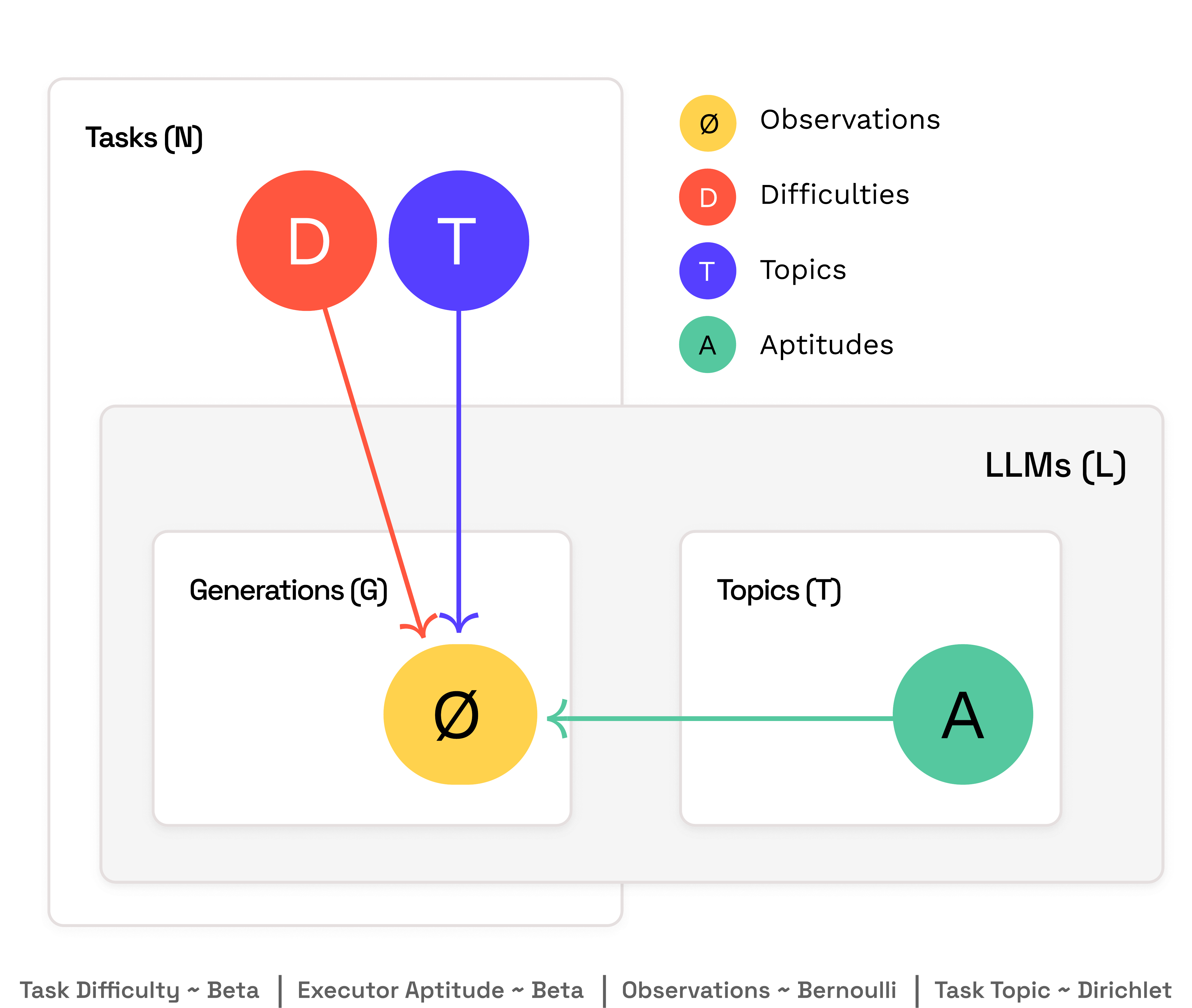}
    \caption{\textbf{Probabilistic graphical model (PGM).} We model an LLM's inherent accuracy, as indirectly observed over generations (G) for topics (T) as a function of the prompt difficulty (D) and the model aptitude (A). We depict this model here in Plate Notation, a standard way of writing the generative process for Bayesian models. $D_n$ induces correlation across models for each prompt.}
    \label{fig:pgm}
\end{figure}

We introduce a generative model~\cite{koller2009probabilistic} for observations $\phi_{nlg}$ (shown in Fig.~\ref{fig:pgm}) that allows direct estimation of true performance parameters. The intuition behind the model is: (1) every prompt has a difficulty $D$, (2) every prompt belongs to a topic $T$, e.g. coding, math, or some weighted combination of them, (3) every LLM has some aptitude $A$ on each topic.
The observed performance of an LLM on a given prompt is a function of the prompt's difficulty, topic combination of the prompt, and the LLM's aptitude on those topics. 

\textbf{Latent variables:}
\begin{itemize}
    \item $D_n \in [0,1]$: Task difficulty for prompt $n$
    \item $T_n \in \{1, \ldots, K\}$: Topic assignment for prompt $n$
    \item $A_{tl} \in [0,1]$: Aptitude of model $l$ on topic $t$
\end{itemize}


\textbf{Generative process:}
\begin{align}
    D_n &\sim \text{Beta}(\alpha_D, \beta_D) \\
    T_n &\sim \text{Categorical}(\boldsymbol{\theta}) \quad \text{where } \boldsymbol{\theta} \sim \text{Dirichlet}(\boldsymbol{\alpha}) \\
    A_{tl} &\sim \text{Beta}(\alpha_{tl}, \beta_{tl}) \\
    \phi_{nlg} &\sim \text{Bernoulli}(\pi_{nl})
\end{align}

\textbf{Link function:}
\begin{equation}\label{eqn:link}
    \pi_{nl} = f(D_n, A_{T_n, l})
\end{equation}
A simple multiplicative form of $(1 - D_n) \cdot A_{T_n, l}$ would capture the intuition that success requires both low difficulty and high model aptitude. However, we find the most accurate results were obtained using a feedforward neural network.

\textbf{Limitations.} The PGM has ad-hoc structural choices which can influence results. Checking alignment against synthetic data with known ground truth across a variety of regimes de-risks these choices.

\textbf{Inference.} We use stochastic variational inference with factorized posterior $q(D_n)q(T_n)\prod_{t,l} q(A_{tl})$. We set uniform priors ($\alpha_D = \beta_D = 1$, $\alpha_{tl} = \beta_{tl} = 1$, $\alpha_t = 1$) and run until convergence.

\textbf{Computing the unbiased oracle:}
\begin{equation}\label{eqn:pgm_oracle}
    \mathcal{O}^{true} = \frac{1}{N} \sum_n \max_l \hat{\pi}_{nl}
\end{equation}
where $\hat{\pi}_{nl}$ are the inferred success probabilities.

\textbf{Independence assumptions.} Fig.~\ref{fig:pgm} assumes conditional independence across generations given latent variables. This is reasonable when temperature-based sampling dominates, but may underestimate correlations when models share training data and architecture.

\subsection{Capability Frontier for Multi-Objective Routing}
\label{ssec:frontier}

Real routing decisions involve multiple objectives. We define the Capability Frontier as the Pareto-optimal surface achievable through routing.

For normalized quality $Q^*$ and cost $C^*$:
\begin{align}
    \phi(\alpha) &= \alpha Q^*_{nlg} + (1-\alpha)(-C^*_{nlg}) \\
    Q^*_{nlg} &= \frac{Q_{nlg} - \min \mathbf{Q}}{\max \mathbf{Q} - \min \mathbf{Q}} \\
    C^*_{nlg} &= \frac{C_{nlg} - \min \mathbf{C}}{\max \mathbf{C} - \min \mathbf{C}}
\end{align}

Sweeping $\alpha \in [0,1]$ traces the Capability Frontier. For debiasing, we:
\begin{enumerate}
    \item Use $\phi(\alpha)$ to determine routing decisions
    \item Apply debiasing separately to quality and cost
\end{enumerate}

For cost (positive real values), we replace the Bernoulli likelihood with LogNormal in our PGM.

\subsection{Posthoc Oracle}
\label{ssec:posthoc}

When a verifier is available at inference time, we can select among multiple generations \textit{after} observing outputs. With $k$ generations per model and a perfect judge:
\begin{equation}\label{eqn:posthoc}
    \mathcal{O}^{kshot}(k) = \frac{1}{N \binom{G}{k}} \sum_n \max_l \sum_{\substack{\mathcal{S} \subseteq [G] \\ |\mathcal{S}| = k}} \max_{j \in \mathcal{S}} \phi_{nlj}
\end{equation}

Using the PGM:
\begin{equation}\label{eqn:posthoc-pgm}
    \mathcal{O}^{kshot}(k) = \frac{1}{N} \sum_n \left[1 - \prod_l (1 - \pi_{nl})^k\right]
\end{equation}


Eqn.~\ref{eqn:posthoc} \& \ref{eqn:posthoc-pgm} formulation is the most naive form of a posthoc router, where all LLMs are queried for every prompt. A tighter bound can be achieved using more efficient posthoc techniques, such as sequential prompting LLMs with a return early rule. This paper does not discuss those approaches but we believe the gains can be attained at lower cost.

\textbf{Critical caveats:}
\begin{itemize}
    \item Assumes a \textit{perfect} judge (zero error)
    \item Assumes the judge is \textit{free} (zero cost)
\end{itemize}

\section{Experimental Setup}
\label{sec:setup}

\paragraph{Benchmarks.}
We evaluate on 16 benchmarks with verifiable correct answers, spanning:
\begin{itemize}
    \item \textbf{Coding:} LiveCodeBench~\cite{jain2024livecodebench}, BigCodeBench~\cite{zhuo2024bigcodebench}, HumanEval-X-Python, HumanEval-X-CPP, HumanEval-X-Javascript, HumanEval-Java, HumanEval-X-Go~\cite{zheng2023humanevalx}, MBPP~\cite{austin2021mbpp}, LeetCode Hard~\cite{leetcodePlatform}
    \item \textbf{Reasoning:} LiveBench-Reasoning~\cite{white2024livebench}, GPQA Diamond~\cite{rein2023gpqa}
    \item \textbf{Instruction-following:} LiveBench-IFEval~\cite{white2024livebench}
    \item \textbf{Medical:} MedCalcBench~\cite{khandekar2024medcalcbnch}
    \item \textbf{Factuality:} TruthfulQA~\cite{lin2022truthfulqa}
    \item \textbf{Agentic:} Terminal-Bench 2.0~\cite{laude2024terminalbench}, LiveCodeBench~\cite{jain2024livecodebench}
\end{itemize}

These benchmarks have binary correctness metrics (pass/fail for code, exact match for QA), enabling clean oracle analysis.

\paragraph{Models.}
We evaluate 21 LLMs spanning major providers:
\begin{itemize}
    \item \textbf{OpenAI:} GPT-5-nano, GPT-5-mini, GPT-5.1~\cite{openai2025gpt51}
    \item \textbf{Anthropic:} Claude Haiku 4.5, Claude Sonnet 4.5~\cite{anthropic2025claude45WhatsNew}
    \item \textbf{Google:} Gemini 2.5 Pro, Gemini 2.5 Flash, Gemini 2.5 Flash-Lite~\cite{googleCloud2025gemini25vertex}
    \item \textbf{Meta:} Llama 4 Scout, Llama 4 Maverick~\cite{hf2025llama4release}
    \item \textbf{Mistral:} Codestral 2508, Devstral Medium 2505, Devstral Small 2505, Mistral Small Instruct~\cite{mistral2025codestral2508}
    \item \textbf{Qwen:} Qwen3 Coder Plus, Qwen3 Coder Flash, Qwen 2.5 Max, Qwen 2.5 72B Instruct~\cite{qwen2025qwen3techreport}
    \item \textbf{Moonshot:} Kimi K2~\cite{moonshot2025kimik2}
    \item \textbf{DeepSeek:} DeepSeek R1~\cite{deepseekai2025deepseekr1}
    \item \textbf{Z.AI:} GLM-4.6~\cite{zai2025glm46v}
\end{itemize}


\paragraph{Generation parameters.}
For all models we use the provider's default hyper-parameters. Where benchmarks have a max tokens specified, we preserve that setting.
\noindent \textbf{Metrics.}
\begin{itemize}
    \item \textbf{Quality:} Accuracy (fraction of prompts answered correctly). For coding benchmarks, we use execution based verification.
    \item \textbf{Cost:} Total API cost in USD (input + output tokens $\times$ provider pricing).
\end{itemize}

\paragraph{Generations.}
Each prompt-model pair evaluated with $G=10$ independent generations. This yields $N \times L \times G$ total observations per benchmark.

\paragraph{Cost measurement.}
Costs computed using provider API pricing as of 01 Jan 2026.

\paragraph{Agentic Benchmarks.}\label{ssec:agentic}

For agentic benchmarks, computing the true oracle is combinatorially hard since the optimal LLM may differ at each trajectory step. To simplify, we fix the LLM within each trajectory. This may \textit{understate} routing benefits; true per-step routing could yield higher gains. We use the mini-SWE-agent~\cite{sweagent2025minisweagent} with default parameters.

\paragraph{Synthetic oracle evaluation (PGM study).}
In addition to real benchmarks, we run a controlled synthetic study to isolate how task heterogeneity drives oracle gains. Data are generated from the probabilistic graphical model defined in Figure \ref{fig:pgm}. We simulate multiple runs of $L=10$ LLMs across $T=30$ latent topics, with $N=1{,}000$ datapoints and $G=10{,}000$ generations per LLM per datapoint, with differing distributions of diversity. Full details can be found in Appendix~\ref{app:entropy}.

Each datapoint is assigned a latent topic drawn from a Dirichlet distribution, whose concentration parameters are varied to sweep from high-entropy topic mixtures (near-uniform) to low-entropy regimes dominated by a single topic. Task difficulty is sampled per datapoint as $D \sim \text{Beta}(1,1)$, yielding a uniform difficulty distribution. Model aptitude is topic-specific, with each model–topic pair assigned an aptitude $A_{l,t} \sim \text{Beta}(5,5)$, inducing moderate specialization without extreme outliers.

For each instantiation of a simulation run, we compute the entropy  and measure oracle performance as the accuracy of the best LLM selected per datapoint. We compare this to the accuracy of the globally best single LLM, reporting oracle uplift as their difference.

\section{Results}
\label{sec:results}

\begingroup
\setlength{\tabcolsep}{3pt}

\begin{table}[t]
\centering
\small

\begin{minipage}[t]{0.44\linewidth}
\centering
\caption{\textbf{Combining LLMs boosts Quality.} Benchmark level breakdown comparing the SOTA LLM with $\mathcal{O}^{true}(\alpha=1)$. Error rate is reduced by 53.7\% on average.}
\label{table:prehoc_quality}

\begin{tabular}{lccc}
\toprule
& \multicolumn{2}{c}{\% Quality} & \multirow{2}{*}{\begin{tabular}[c]{@{}c@{}}\% Error\\ Reduction $\uparrow$\end{tabular}} \\
\cmidrule(lr){2-3}
Benchmark & SOTA & $\mathcal{O}^{\text{true}}$ \\
\midrule
LiveBench-Coding & 82.2 & 86.8 & 26.2 \\
BigCodeBench & 35.8 & 49.1 & 20.6 \\
LeetCode & 79.1 & 87.7 & 41.3 \\
HumanEval-X (Python) & 97.4 & 99.5 & 79.6 \\
HumanEval-X (CPP) & 95.1 & 99.3 & 85.7 \\
HumanEval-X (Javascript) & 93.5 & 97.0 & 53.5 \\
HumanEval-X (Java) & 95.9 & 99.0 & 75.6 \\
HumanEval-X (Go) & 91.3 & 97.1 & 66.0 \\
MBPP & 86.2 & 92.3 & 44.3 \\
MedCalcBench & 70.5 & 86.4 & 53.9 \\
TruthfulQA & 98.8 & 99.5 & 57.1 \\
LiveBench-IFEval & 80.0 & 87.4 & 36.9 \\
LiveBench-Reasoning & 92.4 & 96.2 & 50.2 \\
GPQA Diamond & 94.0 & 99.0 & 82.7 \\
Terminal-Bench 2.0 & 40.8 & 49.0 & 13.9 \\
LiveBench-Coding (agentic) & 85.5 & 96.0 & 72.1 \\
\midrule
\textbf{Average} & 82.4 & 88.8 & 53.7 \\
\bottomrule
\end{tabular}

\end{minipage}
\hfill
\begin{minipage}[t]{0.44\linewidth}
\centering
\caption{\textbf{Combining LLMs reduces cost.} Cost breakdown comparing SOTA with $\mathcal{O}^{true}(\alpha=\alpha^*)$. Total token cost is reduced by 85.2\% on average.}
\label{table:prehoc_cost}

\begin{tabular}{lccc}
\toprule
& \multicolumn{2}{c}{Cost (cents)} & \multirow{2}{*}{\begin{tabular}[c]{@{}c@{}}\% Cost\\ Saving $\uparrow$\end{tabular}} \\
\cmidrule(lr){2-3}
Benchmark & SOTA & $\mathcal{O}^{\text{true}}$ \\
\midrule
LiveBench-Coding & 1.06 & 0.26 & 75.6 \\
BigCodeBench & 0.43 & 0.07 & 84.6 \\
LeetCode & 1.10 & 0.33 & 70.0 \\
HumanEval-X (Python) & 0.32 & 0.02 & 94.8 \\
HumanEval-X (CPP) & 0.21 & 0.02 & 88.3 \\
HumanEval-X (Javascript) & 0.20 & 0.02 & 92.4 \\
HumanEval-X (Java) & 0.21 & 0.03 & 83.5 \\
HumanEval-X (Go) & 0.55 & 0.03 & 94.8 \\
MBPP & 0.14 & 0.01 & 96.3 \\
MedCalcBench & 0.23 & 0.04 & 83.3 \\
TruthfulQA & 0.38 & 0.00 & 99.5 \\
LiveBench-IFEval & 0.27 & 0.02 & 93.1 \\
LiveBench-Reasoning & 1.04 & 0.41 & 60.9 \\
GPQA Diamond & 0.54 & 0.02 & 96.3 \\
Terminal-Bench 2.0 & 260.84 & 25.39 & 90.3 \\
LiveBench-Coding (agentic) & 3.34 & 1.37 & 58.9 \\
\midrule
\textbf{Average} & 16.93 & 1.75 & 85.2 \\
\bottomrule
\end{tabular}

\end{minipage}

\end{table}
\endgroup

\subsection{Finding \#1: LLM Routing Gives Substantial Gains}
\label{ssec:prehoc}

Using debiased oracles, we quantify achievable routing gains (Table~\ref{table:prehoc_quality} \& \ref{table:prehoc_cost}) through computation of the Capability Frontier as described in Sec.~\ref{ssec:frontier} using Eq.~\ref{eqn:extrapolate_smooth}.
We define the SOTA LLM as the model achieving the highest average quality. $\mathcal{O}^{true}(\alpha=1)$ corresponds to the most accurate achievable router, i.e., an oracle selecting the optimal model per query. $\mathcal{O}^{true}(\alpha=\alpha^*)$ denotes the oracle evaluated at the alpha value that matches SOTA quality, capturing the maximal cost savings attainable at equivalent performance.

\textbf{Error rate reduction:} Compared to SOTA LLM, $\mathcal{O}^{true}(\alpha=1)$ achieves a 54\% average error reduction.

\textbf{Cost savings at SOTA quality:} Compared to SOTA LLM, $\mathcal{O}^{true}(\alpha=\alpha^*)$ achieves 85\% average cost savings.

\subsection{Finding \#2: Posthoc Routing Increases Gains}
\label{ssec:posthoc_results}

By leveraging a free and perfect judge at inference time as described in Eq.\ref{eqn:posthoc}, the error rate can be reduced further (Appendix.~\ref{app:posthoc} Tab.~\ref{table:posthoc_k1} \& \ref{table:posthoc_k10}).
The results quantify not only that gain, but how quickly it changes as the number of attempts, $k$, increases from $1 \to 10$.

As described in Sec.~\ref{ssec:posthoc}, this paper uses the most naive form of a posthoc router. We believe these gains can be attained at significantly at lower cost.

\textbf{k = 1:} 66\% error reduction vs SOTA LLM.

\textbf{k=10:} 82\% error reduction vs SOTA LLM.

\subsection{Finding \#3: Naive Oracles Overestimate Gains}
\label{ssec:bias_quant}

We compare $\mathcal{O}^{biased}$ to $\mathcal{O}^{true}$ across benchmarks (App.~\ref{app:bias} Tab.~\ref{table:bias}, Fig.~\ref{fig:bias_quantification}).

\textbf{Quality bias:} Average 1.2\% overestimation.
\textbf{Cost bias:} Average 37.5\% overestimation.

The larger cost bias arises because cost distributions are more skewed, amplifying selection effects.

\subsection{Finding \#4: Model Reliability Varies Substantially}
\label{ssec:reliability}

LLMs with default hyper parameters by design output different responses for the same input when prompted multiple times. An obvious question though, is how consistent are LLMs in solving the problem across these generations. Appendix.~\ref{app:reliability} Tab.~\ref{table:reliability} shows how different LLMs rank for reliability:

\begin{equation}\label{eqn:reliability}
    \text{reliability}(l) = 2 \times \frac{1}{N} \sum_n \left|\bar{\phi}_{nl} - 0.5\right|
\end{equation}

The most reliable LLM we tested was GPT-5-mini with a score of $90.2\%$ and the least reliable was GLM-4.6 at $76.3\%$. There is no significant correlation between Reliability and either Quality or Cost.

\subsection{Finding \#5: Simulations show oracle uplift increases with data diversity.}
\label{sec:datadiversity}

\begin{figure}[H]
\centering
\includegraphics[width=0.7\linewidth]{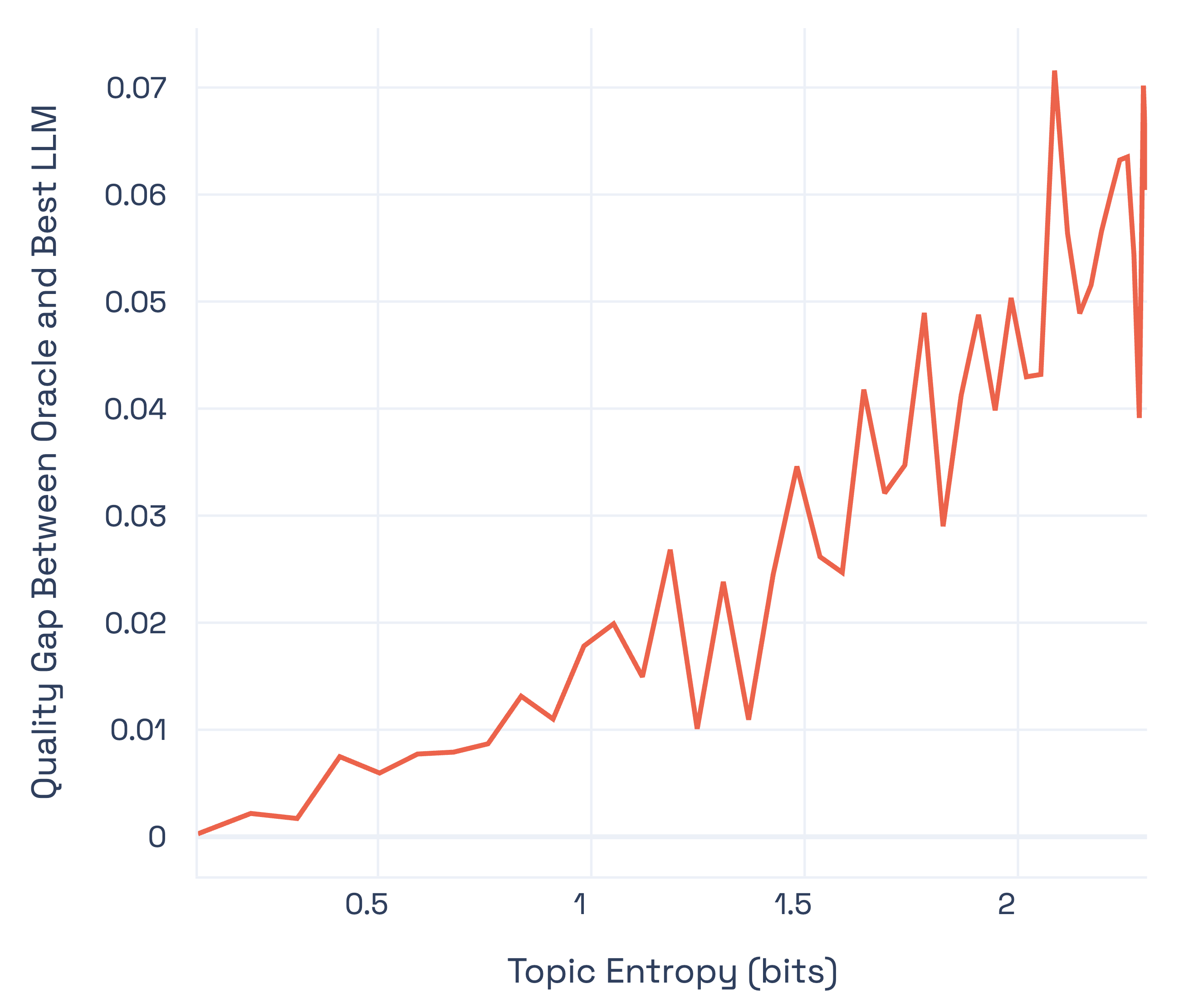}
\caption{Synthetic PGM study measuring the performance gap between an oracle router and best single LLM as task diversity increases. The x-axis shows entropy of the topic distribution, interpolating from single-topic (low entropy) to highly mixed workloads (high entropy). The y-axis reports oracle minus best-single-model accuracy.}
\label{fig:entropy_oracle}
\end{figure}

Figure \ref{fig:entropy_oracle} shows oracle uplift as topic entropy varies in the synthetic PGM study. Uplift increases monotonically with entropy: it is minimal in single-topic regimes and largest under uniform mixtures. This possibly explains why oracle gains may vary across benchmarks and settings, details in App.~\ref{app:entropy}

\section{Limitations}
\label{sec:limitations}

\paragraph{Limited generations.}
With $G=10$, extrapolation carries uncertainty. We mitigate with (1) testing on synthetic data with known ground truth, and (2) comparing to PGM estimates. However larger $G$ would improve estimates.

\paragraph{Agentic Benchmarks Gains May Be Underestimated.}
For agentic benchmarks, computing $\mathcal{O}^{true}$ is combinatorially hard since the optimal LLM may differ at each trajectory step. To simplify, we fix the LLM within each trajectory. This may understate routing benefits; true per-step routing could yield higher gains.

\paragraph{Perfect judge assumption.}
Posthoc oracles assume error-free, cost-free judges. Real verifiers introduce errors and costs that reduce achievable gains.

\section{Conclusion}
\label{sec:conclusion}

This work re-evaluates how the performance of large language models is measured. We show that standard benchmark evaluations, typically based on a single model and a single sampled output per prompt, do not capture the full range of performance that is already attainable with existing models and inference budgets. At the same time, we demonstrate that naive aggregation across models or runs can lead to overly optimistic estimates due to noise. To address both effects, we introduced the \emph{Capability Frontier}, a quality-cost Pareto frontier that characterizes achievable performance while explicitly correcting for these opposing biases.

Empirically, across 21 LLMs and 16 benchmarks, the Capability Frontier substantially outperforms standard single-model evaluations. At matched cost, correcting for single-model evaluation reduces error by 54\% on average, while additionally accounting for single-run variability yields an 82\% reduction. Conversely, at matched accuracy, frontier points often achieve performance comparable to the SOTA LLM at a fraction of the cost. These results suggest that commonly reported benchmark scores can significantly understate achievable system-level performance. Our simulations suggest these gains scale with data heterogeneity: more diverse workloads induce greater model complementarity and larger frontier improvements.

\paragraph{Implications.}
Our findings have several implications for the evaluation and use of LLMs:
\begin{itemize}
    \item \textbf{Evaluation methodology.} Single-model, single-run benchmarks provide a limited view of model capability. Capability Frontier based analysis offers a complementary perspective that accounts for model diversity and sampling effects, and can help contextualize results.
    \item \textbf{System design.} While the Capability Frontier itself is not a deployment strategy, it highlights regimes where simple routing or repeated sampling may be sufficient to achieve large gains, and where more sophisticated methods are necessary to approach the attainable limits.
\end{itemize}

\paragraph{Future work.}
Several extensions are clear. First, incorporating judge error and cost directly into posthoc frontier construction. Second, extending agentic evaluation beyond fixed trajectory routing. Third, developing and evaluating practical routing policies that can approach frontier performance under realistic deployment constraints. Fourthly, studying how system prompt selection and hyper-parameter sampling on LLMs can affect the frontier. Finally, empirically characterizing the link between data diversity and frontier gains remains an important direction.

\newpage
\bibliography{iclr_refs}
\bibliographystyle{icml2026}


\newpage
\appendix

\section{Synthetic Study of Bias Decay}
\label{app:decay}

We studied how Oracle bias decayed for varying numbers of LLMs, correlations between them, and LLM success probabilities. To do this, we leveraged the PGM defined in Sec.~\ref{sssec:pgm} to generate binary tensors of synthetic data. 

For zero LLM correlation, Fig.~\ref{fig:bias-decay} shows results of some of these experiments. As expected, we found that both increasing the number of LLMs, $L$, and LLM success probability, $p$, increased the number of generations, $G$, needed before the bias decayed with $O(G^{-0.5})$. We empirically proved with large enough $G$ this is always the case.

For the regime $L <= 10; 0.3 <= p <= 0.7$, we found that at least $G=50$ generations were needed before Eqn.~\ref{eqn:app-bias} would fit with $c=0.5 \pm0.1$
\begin{equation}\label{eqn:app-bias}
     y = ax^{-b} + c
\end{equation}

When LLM correlation was increased, overall bias reduced along with the number of generations needed before the decay followed a predictable pattern. In the limit of large $G$, through empirical study we found the exponent varied, we found this ranged from $0.25$ to $0.75$ for sensible hyper-parameters. Similar to before, approximately $G=50$ generations were needed for a consistent fit.

\section{Benchmarks}
\label{app:benchmarks}

\begin{figure}[h]
\centering
\includegraphics[width=1\linewidth]{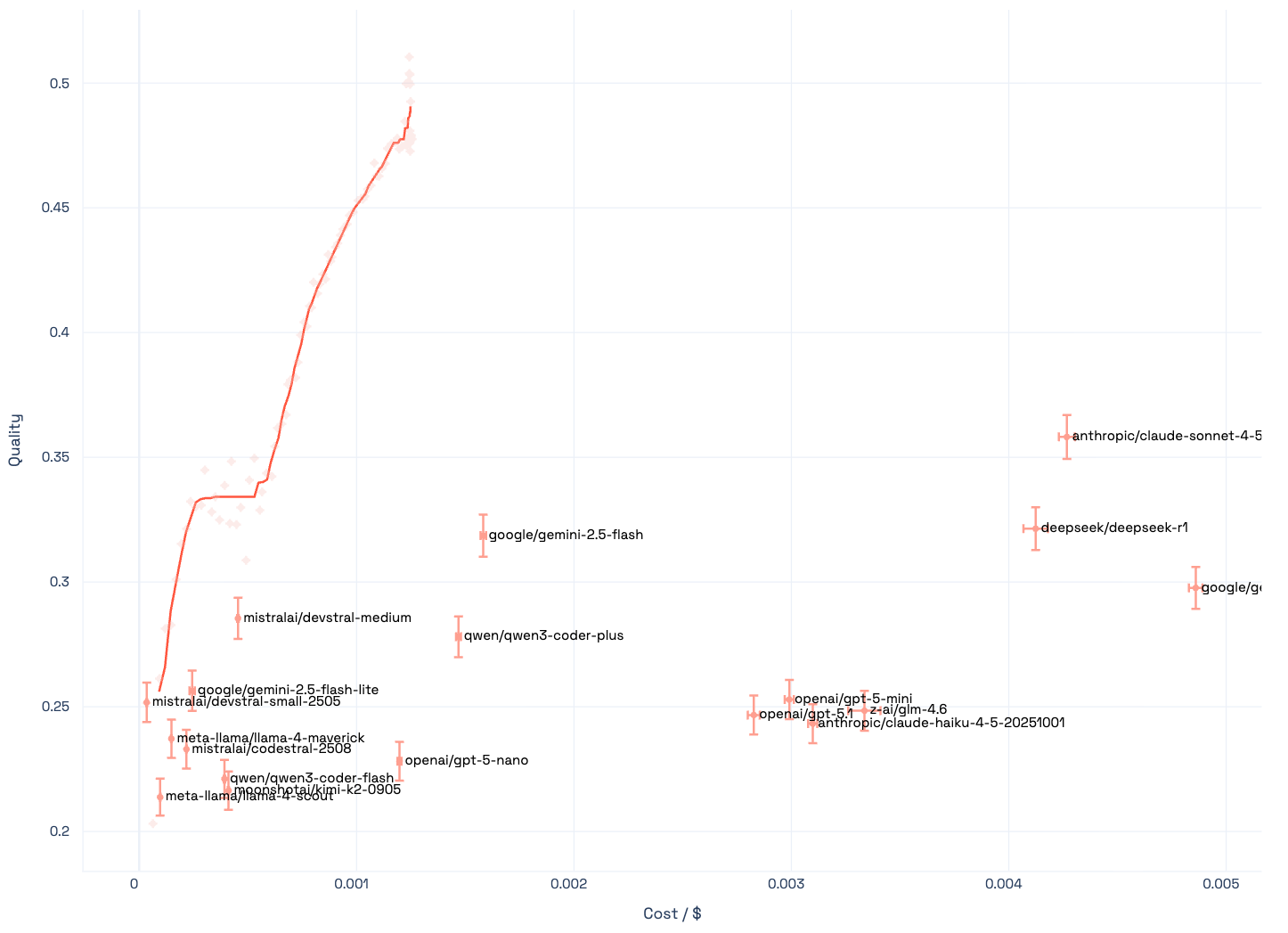}
\caption{BigCodeBench}
\label{fig:bcb}
\end{figure}

Figs.~\ref{fig:bcb}-\ref{fig:term} shows the Capability Frontier for six selected benchmarks. 

\begin{figure}[h]
\centering
\includegraphics[width=1\linewidth]{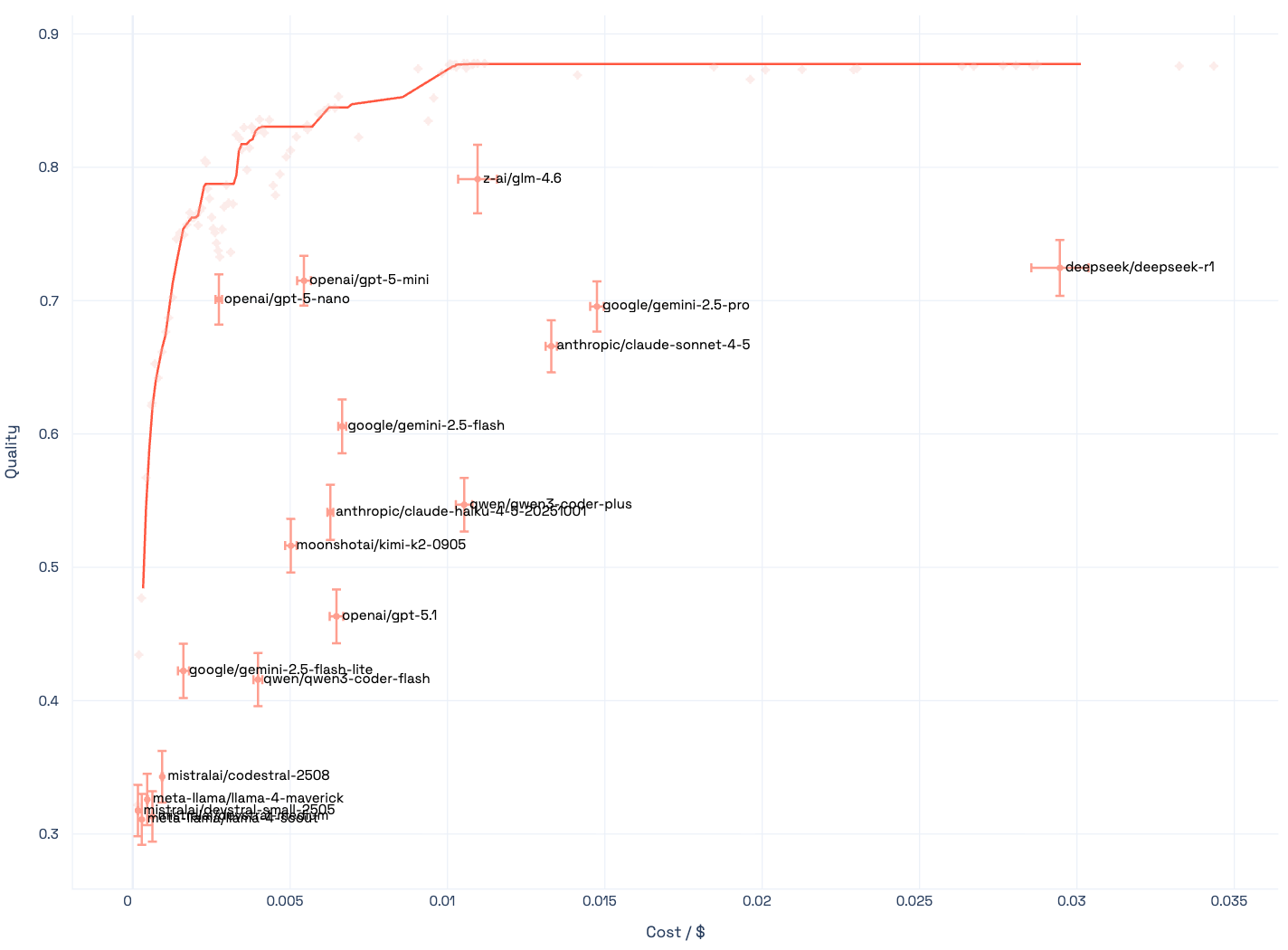}
\caption{LeetCode}
\end{figure}

\begin{figure}[h]
\centering
\includegraphics[width=1\linewidth]{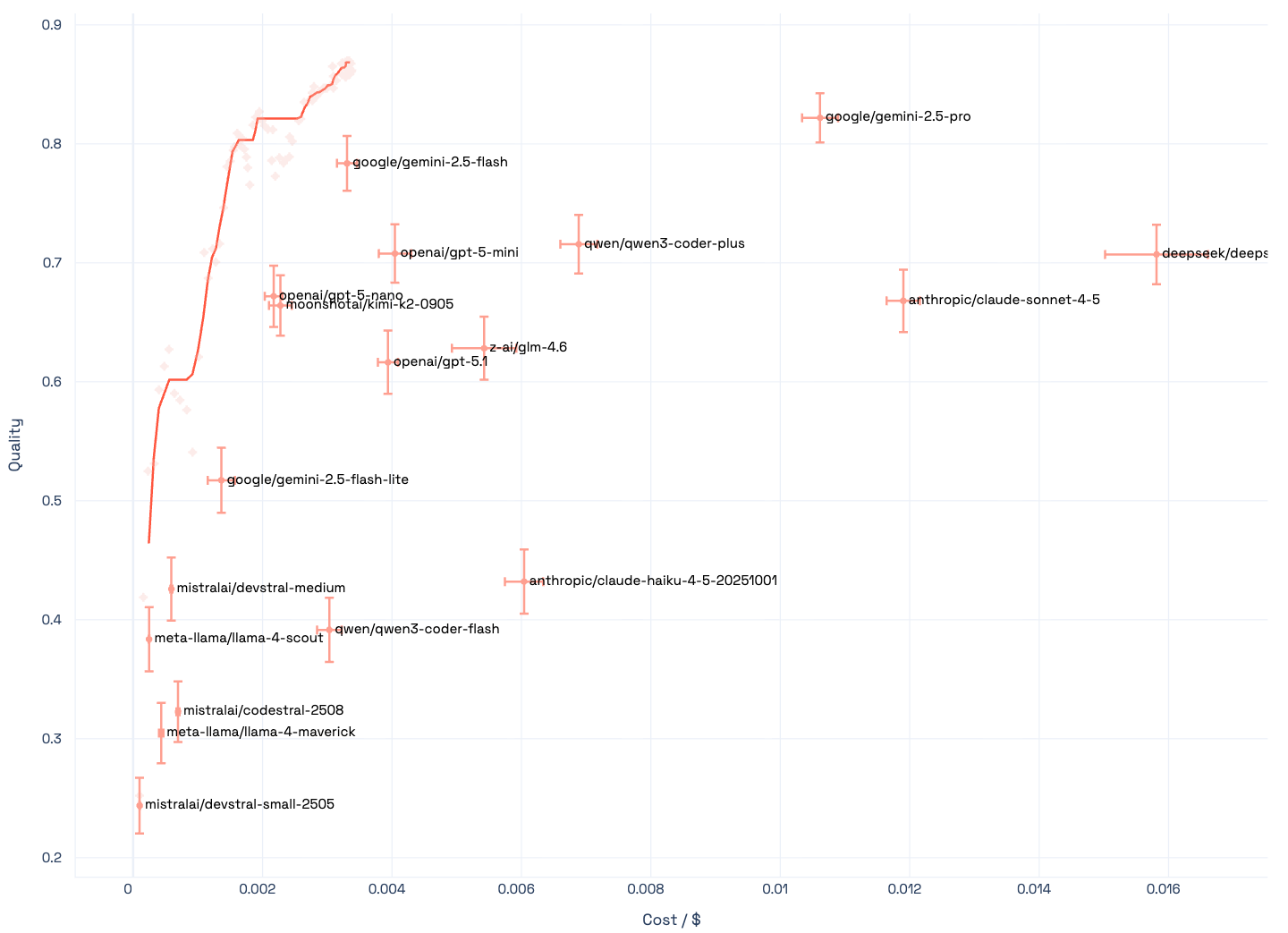}
\caption{LiveBench-Coding}
\end{figure}

\begin{figure}[h]
\centering
\includegraphics[width=1\linewidth]{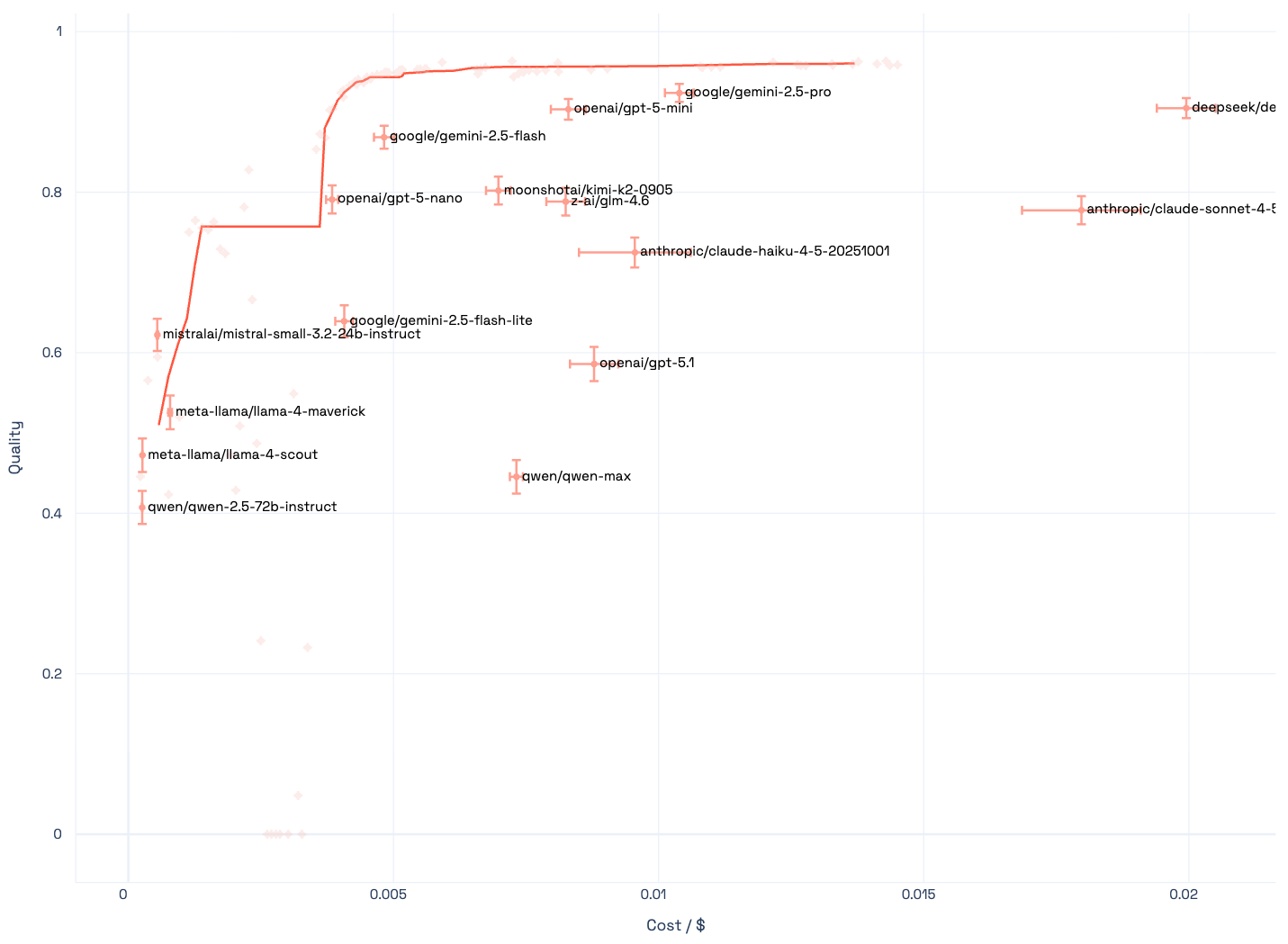}
\caption{LiveBench-Reasoning}
\end{figure}

\begin{figure}[h]
\centering
\includegraphics[width=1\linewidth]{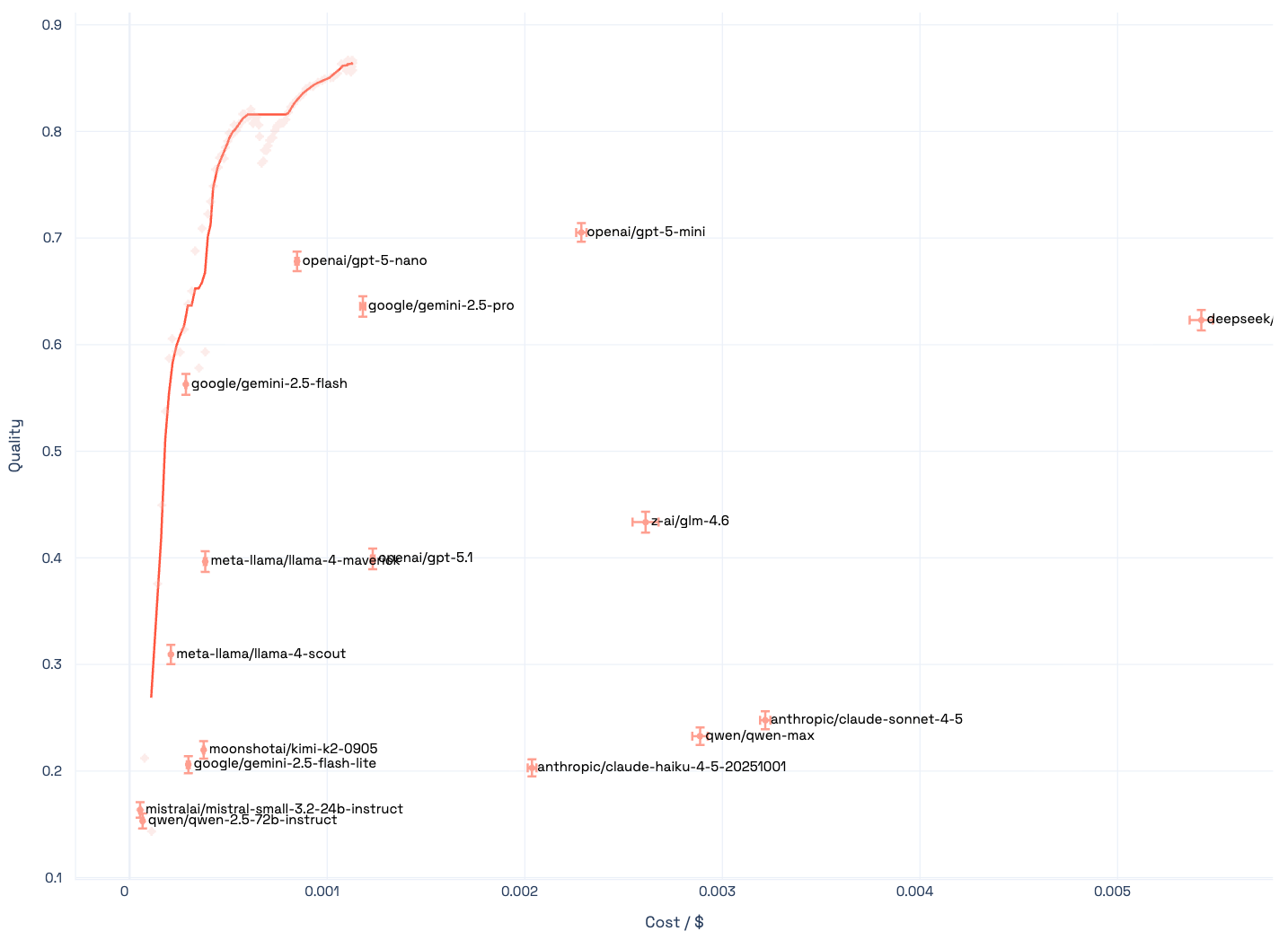}
\caption{MedCalcBench}
\end{figure}

\begin{figure}[h]
\centering
\includegraphics[width=1\linewidth]{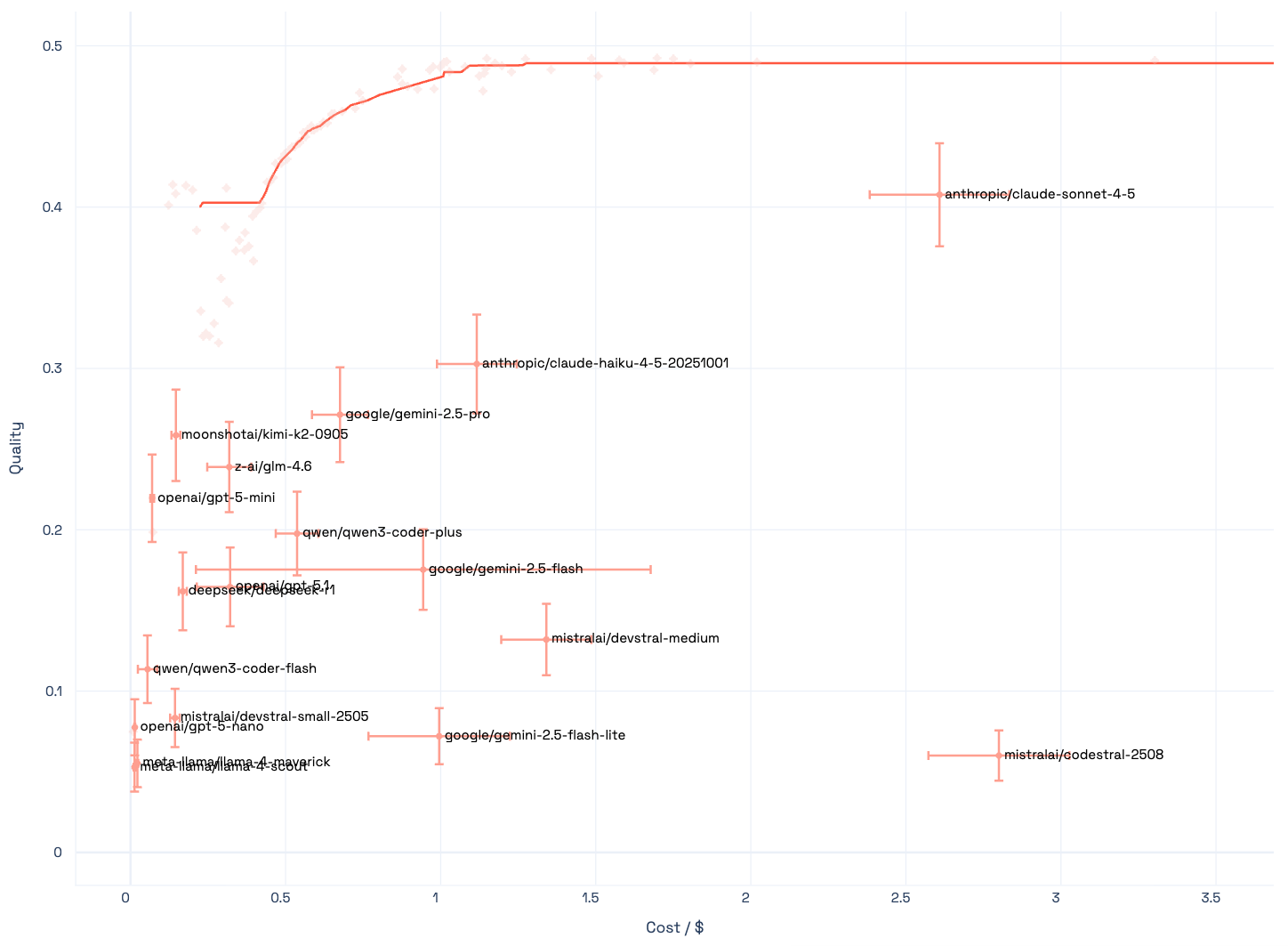}
\caption{Terminal-Bench 2.0 (agentic)}
\label{fig:term}
\end{figure}











\section{LLM Reliability Results}
\label{app:reliability}
Tab.~\ref{table:reliability} summarizes the reliability of LLMs. The scores shown are averaged across all 16 benchmarks.  

\begin{table*}[t]
\centering
\caption{\textbf{LLMs have variable reliability} Comparison of LLMs for reliability in solving a problem. Scores averaged across all benchmarks.}
\label{table:reliability}
\small
\begin{tabular}{lccc}
\toprule
Benchmark & Quality & Cost (\$) & \% Reliability \\
\midrule
\multicolumn{1}{l|}{openai/gpt-5-mini}                        & 0.68                        & 0.0040                   & 90.2               \\
\multicolumn{1}{l|}{anthropic/claude-sonnet-4-5}              & 0.61                        & 0.0615                   & 87.8               \\
\multicolumn{1}{l|}{google/gemini-2.5-pro}                    & 0.69                        & 0.0174                   & 87.6               \\
\multicolumn{1}{l|}{mistralai/codestral-2508}                 & 0.44                        & 0.1308                   & 86.2               \\
\multicolumn{1}{l|}{openai/gpt-5-nano}                        & 0.64                        & 0.0014                   & 85.8               \\
\multicolumn{1}{l|}{mistralai/mistral-small-3.2-24b-instruct} & 0.50                        & 0.0001                   & 85.7               \\
\multicolumn{1}{l|}{anthropic/claude-haiku-4-5-20251001}      & 0.54                        & 0.0256                   & 85.7               \\
\multicolumn{1}{l|}{mistralai/devstral-small}                 & 0.35                        & 0.1158                   & 84.8               \\
\multicolumn{1}{l|}{mistralai/devstral-medium}                & 0.46                        & 0.0549                   & 83.7               \\
\multicolumn{1}{l|}{openai/gpt-5.1}                           & 0.58                        & 0.0098                   & 83.3               \\
\multicolumn{1}{l|}{qwen/qwen-2.5-72b-instruct}               & 0.48                        & 0.0001                   & 83.0               \\
\multicolumn{1}{l|}{deepseek/deepseek-r1}                     & 0.52                        & 0.0104                   & 82.7               \\
\multicolumn{1}{l|}{meta-llama/llama-4-scout}                 & 0.47                        & 0.0005                   & 82.7               \\
\multicolumn{1}{l|}{qwen/qwen3-coder-flash}                   & 0.45                        & 0.0039                   & 82.4               \\
\multicolumn{1}{l|}{google/gemini-2.5-flash-lite}             & 0.52                        & 0.0207                   & 81.7               \\
\multicolumn{1}{l|}{google/gemini-2.5-flash}                  & 0.65                        & 0.0203                   & 81.4               \\
\multicolumn{1}{l|}{qwen/qwen3-coder-plus}                    & 0.53                        & 0.0263                   & 80.8               \\
\multicolumn{1}{l|}{mistralai/devstral-small-2505}            & 0.42                        & 0.0051                   & 80.1               \\
\multicolumn{1}{l|}{qwen/qwen-max}                            & 0.53                        & 0.0023                   & 79.9               \\
\multicolumn{1}{l|}{meta-llama/llama-4-maverick}              & 0.51                        & 0.0008                   & 79.6               \\
\multicolumn{1}{l|}{moonshotai/kimi-k2-0905}                  & 0.51                        & 0.0043                   & 77.1               \\
\multicolumn{1}{l|}{z-ai/glm-4.6}                             & 0.45                        & 0.0098                   & 76.3               \\
\midrule
\textbf{Average} & 0.52 &	0.0239 & 83.1\% \\
\bottomrule
\end{tabular}
\end{table*}

\section{More routing methods}
\label{app:more_routing}
\paragraph{Training-Free and Online Methods}
To support high-volume serving without extensive labeled data, several training-free approaches have emerged. \citet{wu2025efficienttrainingfreeonlinerouting} proposed an online routing mechanism using approximate nearest neighbor search to estimate query features with theoretical performance guarantees. Building on this, CSCR \citep{shirkavand2025cscr} employs cost-aware contrastive learning to map prompts and models into a shared embedding space, facilitating low-latency routing sensitive to both cost and quality. For environments lacking ground-truth labels, Smoothie \citep{guha2024smoothie} provides a label-free framework that leverages weak supervision and model ensembles to estimate query-specific quality.

\paragraph{Cascades and Multi-Objective Optimization}
Routing is often implemented as a cascade, where simpler models are queried before deferring to more expensive ones. FrugalGPT \citep{chen2023frugalgptuselargelanguage} pioneered this via learned cascades to reduce spend without degrading quality. More recently, C3PO \citep{valkanas2025c3po} achieved cost-controlled cascades using conformal prediction to provide provable coverage bounds, while \citet{dekoninck2025unifiedapproachroutingcascading} derived optimal stopping rules for sequential model invocation. Other multi-objective systems, such as xRouter \citep{qian2025xrouter}, utilize reinforcement learning with explicit monetary rewards to navigate the quality-cost trade-off surface.

\paragraph{Agreement-Based and Preference Routing}
Alternative signals for routing include model agreement and human preferences. ABC \citep{kolawole2025abc} uses agreement between models to make deferral decisions, while BEST-Route \citep{ding2025bestroute} jointly optimizes model selection and best-of-n sampling. In the absence of objective correctness, RouteLLM \citep{ong2025routellmlearningroutellms} trains routers on ``Chatbot Arena" style preference data to maintain quality at significantly reduced costs. To improve transparency, LLMRank \citep{agrawal2025llmrank} analyzes specific reasoning patterns to provide a granular understanding of model utility beyond aggregate scores.

\section{Posthoc Routing Gains}
\label{app:posthoc}

Tab.\ref{table:posthoc_k1} and Tab.\ref{table:posthoc_k10} shows the error rate reductions for $k=1$ shot and $k=10$ shot posthoc routing. The average error rate reduction is 66.1\% and 82.4\% respectively. 

\begin{table*}[t]
\centering
\caption{Benchmark level breakdown comparing SOTA LLM with $\mathcal{O}^{kshot}(k=1)$.}
\small
\begin{tabular}{lccccc}
\toprule
& \multicolumn{2}{c}{\% Quality} & \multicolumn{2}{c}{Cost (\$)} &   \multirow{2}{*}{\begin{tabular}[c]{@{}c@{}}\% Error Rate\\ Reduction $\uparrow$ \end{tabular}} \\
\cmidrule(lr){2-3} \cmidrule(lr){4-5}
Benchmark & SOTA LLM & $\mathcal{O}^{\text{kshot}}(1)$ & SOTA LLM & $\mathcal{O}^{\text{kshot}}(1)$ \\
\midrule

LiveBench-Coding             & 82.2  & 89.8  & 0.011 & 0.079                & 43.0                 \\
BigCodeBench                 & 35.8  & 66.7  & 0.004 & 0.032                & 48.1                 \\
LeetCode                     & 79.1  & 89.3  & 0.011 & 0.129                & 48.8                 \\
HumanEval-X (Python)         & 97.4  & 99.6  & 0.003 & 0.029                & 83.7                 \\
HumanEval-X (CPP)            & 95.1  & 99.5  & 0.002 & 0.029                & 88.8                 \\
HumanEval-X (Javascript)     & 93.5  & 97.2  & 0.002 & 0.028                & 57.0                 \\
HumanEval-X (Java)           & 95.9  & 99.4  & 0.002 & 0.036                & 85.1                 \\
HumanEval-X (Go)             & 91.3  & 97.7  & 0.006 & 0.034                & 73.9                 \\
MBPP                         & 86.2  & 93.4  & 0.001 & 0.029                & 51.8                 \\
MedCalcBench                 & 70.5  & 90.8  & 0.002 & 0.023                & 68.6                 \\
TruthfulQA                   & 98.8  & 99.6  & 0.004 & 0.009                & 66.3                 \\
LiveBench-IFEval             & 80.0  & 92.7  & 0.003 & 0.027                & 63.4                 \\
LiveBench-Reasoning          & 92.4  & 97.8  & 0.010 & 0.112                & 70.7                 \\
GPQA Diamond                 & 94.0  & 100.0 & 0.005 & 0.036                & 100.0                \\
Terminal-Bench 2.0 (agentic) & 40.8  & 57.4  & 2.608 & 12.289               & 28.1                 \\
LiveBench-Coding (agentic)   & 85.5  & 97.1  & 0.033 & 2.017                & 79.7                 \\

\midrule
\textbf{Average} & 82.4  & 91.7  & 0.169 & 0.934                & 66.1 \\
\bottomrule
\end{tabular}
\label{table:posthoc_k1}
\end{table*}

\begin{table*}[t]
\centering
\caption{Benchmark level breakdown comparing SOTA LLM with $\mathcal{O}^{kshot}(k=10)$.}
\small
\begin{tabular}{lccccc}
\toprule
& \multicolumn{2}{c}{\% Quality} & \multicolumn{2}{c}{Cost (\$)} &   \multirow{2}{*}{\begin{tabular}[c]{@{}c@{}}\% Error Rate\\ Reduction $\uparrow$ \end{tabular}} \\
\cmidrule(lr){2-3} \cmidrule(lr){4-5}
Benchmark & SOTA LLM & $\mathcal{O}^{\text{kshot}}(10)$ & SOTA LLM & $\mathcal{O}^{\text{kshot}}(10)$ \\
\midrule

LiveBench-Coding             & 82.2  & 93.8  & 0.011 & 0.789                & 64.9                 \\
BigCodeBench                 & 35.8  & 77.5  & 0.004 & 0.318                & 65.0                 \\
LeetCode                     & 79.1  & 93.9  & 0.011 & 1.292                & 70.6                 \\
HumanEval-X (Python)         & 97.4  & 100.0 & 0.003 & 0.290                & 100.0                \\
HumanEval-X (CPP)            & 95.1  & 100.0 & 0.002 & 0.288                & 100.0                \\
HumanEval-X (Javascript)     & 93.5  & 98.2  & 0.002 & 0.275                & 72.0                 \\
HumanEval-X (Java)           & 95.9  & 100.0 & 0.002 & 0.361                & 100.0                \\
HumanEval-X (Go)             & 91.3  & 98.2  & 0.006 & 0.339                & 78.9                 \\
MBPP                         & 86.2  & 95.5  & 0.001 & 0.286                & 67.3                 \\
MedCalcBench                 & 70.5  & 96.6  & 0.002 & 0.234                & 88.5                 \\
TruthfulQA                   & 98.8  & 99.9  & 0.004 & 0.088                & 89.1                 \\
LiveBench-IFEval             & 80.0  & 100.0 & 0.003 & 0.269                & 100.0                \\
LiveBench-Reasoning          & 92.4  & 98.5  & 0.010 & 1.121                & 80.4                 \\
GPQA Diamond                 & 94.0  & 100.0 & 0.005 & 0.358                & 100.0                \\
Terminal-Bench 2.0 (agentic) & 40.8  & 74.2  & 2.608 & 122.891              & 56.4                 \\
LiveBench-Coding (agentic)   & 85.5  & 97.8  & 0.033 & 20.166               & 85.1                 \\

\midrule
\textbf{Average} & 82.4  & 95.2  & 0.169 & 9.335                & 82.4 \\
\bottomrule
\end{tabular}
\label{table:posthoc_k10}
\end{table*}

\section{Quantifying the Bias}
\label{app:bias}
\begin{figure*}[h]
\centering
\begin{tikzpicture}

\begin{axis}[
    name=quality,
    width=0.48\textwidth,
    height=0.85\textwidth,
    xlabel={Quality Decrease (\%)},
    xmin=0, xmax=9.5,
    ytick=data,
    yticklabels={
        LiveBench-Coding (agentic),
        Terminal-Bench 2.0 (agentic),
        GPQA Diamond,
        LiveBench-Reasoning,
        LiveBench-IFEval,
        TruthfulQA,
        MedCalcBench,
        MBPP,
        HumanEval-X (Go),
        HumanEval-X (Javas),
        HumanEval-X (Javascript),
        HumanEval-X (CPP),
        HumanEval-X (Python),
        LeetCode,
        BigCodeBench,
        LiveBench-Coding
    },
    yticklabel style={font=\small},
    grid=major,
    grid style={dashed, gray!30},
    ymajorgrids=true,
    xmajorgrids=true,
    enlarge y limits=0.08,
    clip=false
]

\draw[blue, dotted, thick] (axis cs:1.0,0) -- (axis cs:1.0,15.5);

\addplot[only marks, mark=*, mark size=3pt, blue] coordinates {
    (0.051, 0)    
    (2.951, 1)    
    (0.034, 2)    
    (0.524, 3)    
    (0.888, 4)    
    (0.022, 5)    
    (0.971, 6)    
    (0.208, 7)    
    (0.081, 8)    
    (0.082, 9)    
    (0.109, 10)    
    (0.027, 11)    
    (0.048, 12)    
    (0.16, 13)    
    (9.246, 14)    
    (0.569, 15)    
    };

\end{axis}

\begin{axis}[
    at={(quality.east)},
    anchor=west,
    xshift=0.8cm,
    width=0.48\textwidth,
    height=0.85\textwidth,
    xlabel={Cost Decrease (\%)},
    xmin=0, xmax=95,
    ytick=data,
    yticklabels={},  
    grid=major,
    grid style={dashed, gray!30},
    ymajorgrids=true,
    xmajorgrids=true,
    enlarge y limits=0.08,
    clip=false
]

\draw[red, dotted, thick] (axis cs:23.8,0) -- (axis cs:23.8,15.5);

\addplot[only marks, mark=triangle*, mark size=4pt, red] coordinates {
    (11.98, 0)    
    (34.98, 1)    
    (3.53, 2)    
    (56.3, 3)    
    (9.81, 4)    
    (1.47, 5)    
    (10.6, 6)    
    (43.58, 7)    
    (12.24, 8)    
    (68.04, 9)    
    (24.11, 10)    
    (32.56, 11)    
    (30.75, 12)    
    (30.17, 13)    
    (3.77, 14)    
    (7.21, 15)    
    };

\end{axis}

\end{tikzpicture}
\caption{Bias quantification across benchmarks. The biased oracle $\mathcal{O}^{biased}$ overestimates both Quality (left, blue circles) and Cost (right, red triangles) relative to the true oracle $\mathcal{O}^{true}$. Quality bias is modest (under 9\%, averaging $\approx 1.0$\%), while cost bias varies substantially (1.5--68\%, averaging $\approx 23.8$\%). Dotted vertical lines indicate averages. See Tab.\ref{table:bias} for detailed values.}
\label{fig:bias_quantification}
\end{figure*}
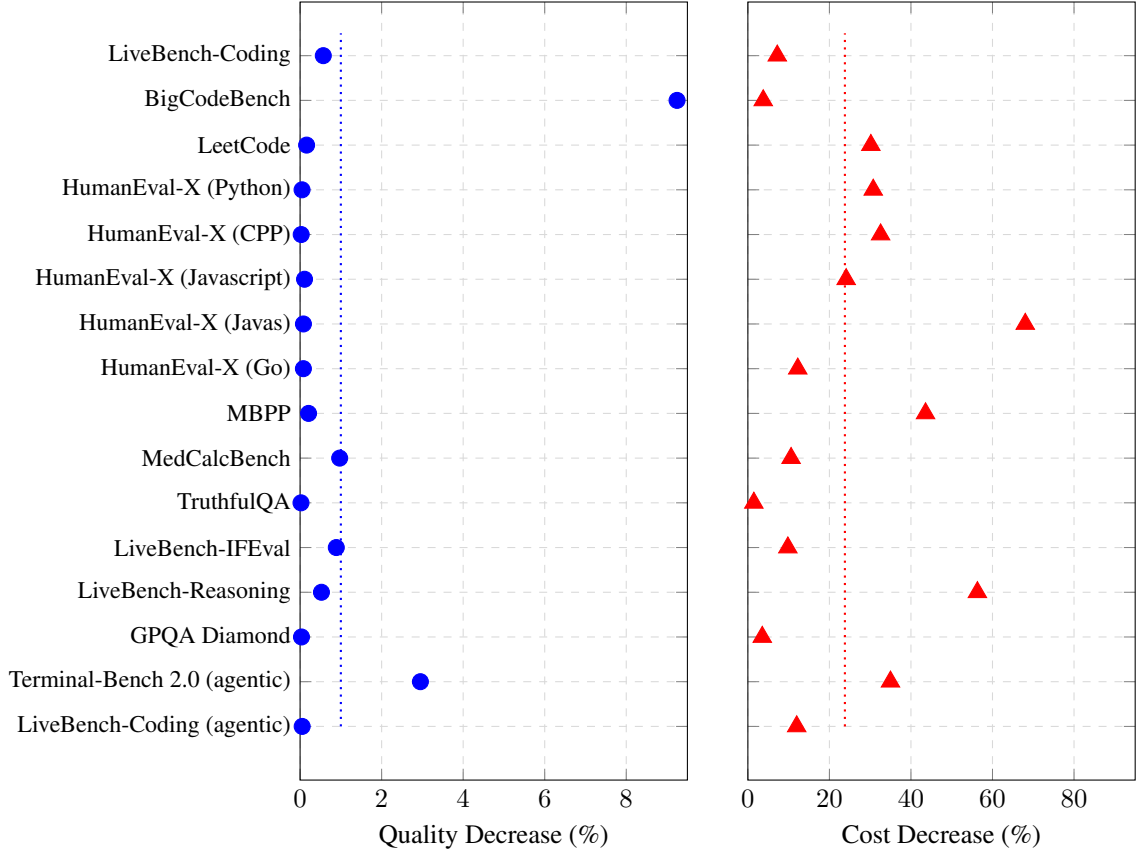

Fig.\ref{fig:bias_quantification} shows the bias quantification across 16 benchmarks.

\begin{table*}[t]
\centering
\caption{\textbf{Naive oracles systematically overestimate.} Comparison of biased and debiased oracle estimates. Cost bias is substantially larger than quality bias.}
\label{table:bias}
\small
\begin{tabular}{lcccc}
\toprule
& \multicolumn{2}{c}{Quality} & \multicolumn{2}{c}{Cost} \\
\cmidrule(lr){2-3} \cmidrule(lr){4-5}
Benchmark & $\mathcal{O}^{\text{biased}}$ & $\mathcal{O}^{\text{true}}$ & $\mathcal{O}^{\text{biased}}$ & $\mathcal{O}^{\text{true}}$ \\
\midrule
LiveBench-Coding             & 87.3            & 86.8           & 0.31          & 0.33          \\
BigCodeBench                 & 53.6            & 49.1           & 0.12          & 0.12          \\
LeetCode                     & 87.9            & 87.7           & 0.74          & 1.05          \\
HumanEval-X (Python)         & 99.5            & 99.5           & 0.02          & 0.02          \\
HumanEval-X (CPP)            & 99.3            & 99.3           & 0.03          & 0.05          \\
HumanEval-X (Javascript)     & 97.1            & 97.0           & 0.02          & 0.03          \\
HumanEval-X (Java)           & 99.1            & 99.0           & 0.05          & 0.15          \\
HumanEval-X (Go)             & 97.1            & 97.1           & 0.05          & 0.06          \\
MBPP                         & 92.5            & 92.3           & 0.03          & 0.06          \\
MedCalcBench                 & 87.3            & 86.5           & 0.10          & 0.11          \\
TruthfulQA                   & 99.5            & 99.5           & 0.01          & 0.01          \\
LiveBench-IFEval             & 88.1            & 87.3           & 0.12          & 0.13          \\
LiveBench-Reasoning          & 96.6            & 96.0           & 0.59          & 1.35          \\
GPQA Diamond                 & 99.0            & 99.0           & 0.02          & 0.02          \\
Terminal-Bench 2.0 (agentic) & 50.4            & 48.9           & 82.93         & 127.55        \\
LiveBench-Coding (agentic)   & 96.0            & 96.0           & 1.55          & 1.76          \\ 
\midrule
\textbf{Average} & 89.4            & 88.8           & 5.42          & 8.30 \\
\bottomrule
\end{tabular}
\end{table*}

\section{Topic Entropy Study}\label{app:entropy}
We ran a Synthetic study measuring the affect of topic distribution entropy on Oracle uplift, defined as:
\begin{equation}
    uplift = \mathcal{O}^{true} - \max_l \frac{1}{N} \sum_n \bar \phi_{nl}
\end{equation}

10 million observations (1000 datapoints, 10000 generations) were generated using the PGM defined in Sec.~\ref{sssec:pgm}. It was configured with:
\begin{enumerate}
    \item 30 latent topics, T
    \item 10 LLMs, K
    \item Task Difficulty Distribution, $D_n \sim Beta(alpha=1, beta=1)\ i.e.\ uniform$
    \item LLM Topic Aptitude, $A_{tl} \sim Beta(alpha=5, beta=5)$
\end{enumerate}

We linearly varied Topic Distribution, $T_n \sim Diriclet(\boldsymbol \alpha)$, from uniform $\alpha_k = \frac{1}{K}$ (high entropy), to a single topic $\alpha_1=1,\ \alpha_{2:K}=0$ (low entropy). Topic entropy given by:
\begin{equation}
    H(\boldsymbol{\alpha}) = -\sum_{k=1}^K \alpha_k \log \alpha_k
\end{equation}

\section{LLM usage}\label{app:llm_usage}
We used large language models (LLMs) solely for light writing and editorial assistance. Specifically, LLMs were used to suggest minor improvements to grammar, clarity, and flow in portions of the manuscript. All technical contributions, empirical findings, and conclusions are the original work of the authors. We reviewed and verified all LLM-assisted edits to ensure accuracy and alignment with the intended meaning.

\end{document}